\documentclass[letterpaper]{article} 
\usepackage{aaai24}  
\usepackage{times}  
\usepackage{helvet}  
\usepackage{courier}  
\usepackage[hyphens]{url}  
\usepackage{graphicx} 
\def\UrlFont{\rm}  
\usepackage{natbib}  
\usepackage{caption} 
\usepackage{algorithm}
\usepackage{algorithmic}

\usepackage{newfloat}
\usepackage{listings}
\usepackage{amsmath}
\usepackage{makecell}
\usepackage{multirow}
\usepackage{amssymb}
\usepackage{bm}
\usepackage{xcolor}         
\usepackage{multirow}
\usepackage{booktabs}
\usepackage{subcaption}
\usepackage[hang,multiple,flushmargin]{footmisc}
\usepackage{hyperref}
\DeclareCaptionStyle{ruled}{labelfont=normalfont,labelsep=colon,strut=off} 
\floatstyle{ruled}
\newfloat{listing}{tb}{lst}{}
\floatname{listing}{Listing}
\title{Regulating Intermediate 3D Features for Vision-Centric Autonomous Driving}
\author{
    Junkai Xu\textsuperscript{\rm 1,2\thanks{Work performed during an internship at FABU Inc.}}\quad Liang Peng\textsuperscript{\rm 1,2,*}\quad Haoran Cheng\textsuperscript{\rm 1,2,*} \quad Linxuan Xia\textsuperscript{\rm 1,2,*} \\ Qi Zhou\textsuperscript{\rm 1,2,*} \quad Dan Deng\textsuperscript{\rm 2} \quad Wei Qian\textsuperscript{\rm 2} \quad Wenxiao Wang\textsuperscript{\rm 3\thanks{Corresponding author}} \quad Deng Cai\textsuperscript{\rm 1,2} \\
{\tt\small \{xujunkai, pengliang, haorancheng\}@zju.edu.cn} \\
}
\affiliations{
    \textsuperscript{\rm 1}State Key Lab of CAD \& CG, Zhejiang University \quad
    \textsuperscript{\rm 2}FABU Inc. \\
    \textsuperscript{\rm 3}School of Software Technology, Zhejiang University \\

}

\usepackage{bibentry}

\begin{document}

\maketitle

\begin{abstract}

    Multi-camera perception tasks have gained significant attention in the field of autonomous driving.
    However, existing frameworks based on Lift-Splat-Shoot (LSS) in the multi-camera setting cannot produce suitable dense 3D features due to the projection nature and uncontrollable densification process.
    To resolve this problem, we propose to regulate intermediate dense 3D features with the help of volume rendering.
    Specifically, we employ volume rendering to process the dense 3D features to obtain corresponding 2D features (\textit{e.g.,} depth maps, semantic maps), which are supervised by associated labels in the training.
    This manner regulates the generation of dense 3D features on the feature level, providing appropriate dense and unified features for multiple perception tasks.
    Therefore, our approach is termed \textbf{Vampire}, stands for ``\textbf{V}olume rendering \textbf{A}s \textbf{M}ulti-camera \textbf{P}erception \textbf{I}ntermediate feature \textbf{RE}gulator''.
    Experimental results on the Occ3D and nuScenes datasets demonstrate that Vampire facilitates fine-grained and appropriate extraction of dense 3D features, and is competitive with existing SOTA methods across diverse downstream perception tasks like 3D occupancy prediction, LiDAR segmentation and 3D objection detection, while utilizing moderate GPU resources. 
    We provide a video demonstration in the supplementary materials and Codes are available at \url{github.com/cskkxjk/Vampire}.

\end{abstract}

\section{Introduction}
    Vision-centric 3D surrounding perception plays an important role in modern autonomous driving and robotics due to its convenience and board applicability for downstream tasks. 
    Vision-based perception frameworks can be broadly categorized into two paradigms~\cite{li2023fb}: backward projection (or Transformer-based~\cite{li2022bevstereo}) and forward projection (or LSS-based, as it originates from the concept of ``Lift, Splat, Shoot'' \cite{philion2020lift}).

    Backward projection / Transformer-based approaches set 3D points in 3D space or BEV plane and then projects these points back onto the 2D image. 
    This procedure allows each predefined 3D or BEV position to obtain corresponding image features.
    Transformer \cite{vaswani2017attention} architectures are widely used in this paradigm to aggregate information from image features, generating task-specific features tailored to meet the objectives~\cite{li2022bevformer,huang2023tri,wei2023surroundocc,zhang2023occformer,zhou2022cross,liu2022petr,liu2022petrv2}.
    When demonstrating promising performance on various perception tasks such as 3D object detection \cite{li2022bevformer,liu2022petr}, BEV map segmentation \cite{li2022bevformer,liu2022petrv2} and 3D occupancy prediction \cite{huang2023tri,wei2023surroundocc,zhang2023occformer}, they require substantial GPU memory to support the interaction between task queries and image features \cite{zhang2022beverse}. 
\begin{figure}
  \centering
  \includegraphics[trim=5 8 2 2, clip, width=1.0\linewidth]{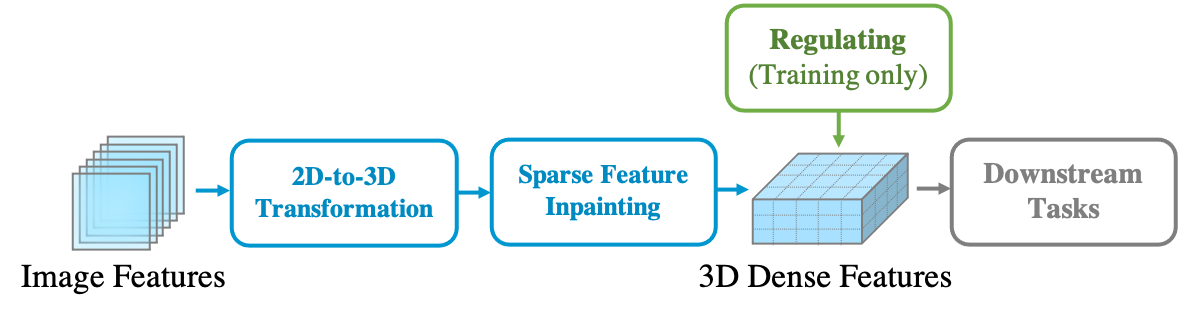}
  \caption{\textbf{Method overview.}
  The key idea is to regulate dense intermediate 3D features in the training stage, to produce appropriate and unified features for different downstream perception tasks.}
  \vspace{-4mm}
  \label{fig:1}
\end{figure}

    In contrast, forward projection / LSS-based methods project 2D image features onto the 3D space, incorporating per-pixel depth estimation.
    They rely on implicit \cite{philion2020lift,hu2021fiery} or explicit \cite{li2022bevdepth,li2022bevstereo,huang2021bevdet} depth estimation to elevate image features to the 3D space, acquiring intermediate feature representations such as BEV or 3D voxel representation for task-specific heads.
    This paradigm is effective for object-level perception tasks, \textit{e.g.}, 3D object detection, but struggles with dense point / grid-level perception tasks, \textit{e.g.}, 3D occupancy prediction. 
    Using estimated per-pixel depth and camera calibrations, these methods position 2D features at the foremost visible surface of objects in the 3D space, leading to sparse 3D features.

    An intuitive resolution would be to densify the sparse 3D features using a feature inpainting module.
    This enables the model to guess and inpaint the empty regions based on known sparse features and produce dense 3D features. 
    However, this naive manner is uncontrollable due to the lack of regulations, and could cause some kind of ``overgeneration'', which means that features may be generated at the wrong places and violate the geometry constraints. 
    Here the question arises: \textit{\textbf{how to find decent regulations in favor of appropriate dense 3D feature extraction?}}
    
    We resort to employing occupancy to model the 3D feature space in a unified manner.
    Occupancy is an ideal dense 3D representation due to its fine-grained information and universality in different tasks~\cite{huang2023tri,tian2023occ3d,sima2023_occnet}.
    We observe that there is an analogy between the occupancy and the volume density in the implicit scene representation NeRF \cite{mildenhall2021nerf,barron2021mip}, as they both describe whether a space is occupied.
    This observation motivates us to employ additional information from the 2D space to implicitly regulate our intermediate 3D features, as NeRF does.

    To this end, we incorporate volume rendering \cite{max1995optical} as a regulator for the intermediate 3D feature space (see Figure \ref{fig:1}). 
    Specifically, we map image features to 3D voxel space following the LSS scheme \cite{philion2020lift}, and employ a 3D hourglass-like design \cite{chang2018pyramid} as the sparse feature inpaintor.
    The resulting dense intermediate 3D features are then used to generate feature volumes (density, semantic) for volume rendering.
    We supervise the rendered depth maps and semantic maps with LiDAR projected ground-truth labels under both camera views and bird's-eye-view.
    In this way, we employ simple 2D supervisions to regulate dense intermediate 3D features, which ensures our sparse feature inpaintor not to generate unreasonable 3D features that could violate their 2D correspondences.
    We term the entire framework as \textbf{Vampire}, stands for taking volume rendering as a regulator for intermediate features in multi-camera perception. 
    We provide the overview design in Figure \ref{fig:1}.
    
    We perform experiments on various multi-camera perception tasks, including 3D occupancy prediction~\cite{tian2023occ3d}, image-based LiDAR segmentation on the competitive nuScenes dataset \cite{caesar2020nuscenes}, and we also assess whether the regulated 3D features continue to exhibit effectiveness for the 3D object detection.
    
    The contributions of this work are summarized as follows:
    
    $\bullet$ We provide a new outlook on intermediate features for vision-centric perception tasks, drawing connections between the occupancy in autonomous driving and volume density in NeRF.
    
    $\bullet$ We introduce Vampire, a multi-camera perception framework. 
    The key component lies in using volume rendering as a regulator for dense intermediate 3D features. 
    As such, different perception tasks benefit from the regulated intermediate features.
    
    $\bullet$ We demonstrate that our method can handle several perception tasks in a single forward pass with moderate computational resources. 
    The single Vampire model that consumes limited GPU memory (12GB per device) for training is comparable with other existing SOTAs across multiple perception tasks (3D occupancy prediction, image-based LiDAR segmentation and 3D objection detection).
 
\section{Related Work}
    \subsection{Multi-camera 3D Perception} 
        3D object detection is a classic and longstanding 3D perception task.
        In multi-camera setting, various attempts \cite{philion2020lift,li2022bevformer,huang2021bevdet,li2022bevdepth} have been proposed for detecting objects in the bird's-eye-view (BEV) representations which collapse the height dimension of 3D space to achieve a balance between accuracy and efficiency.
        LSS \cite{philion2020lift} and its follow-ups \cite{huang2021bevdet,li2022bevdepth,li2022bevstereo} first estimate implicit or explicit per-pixel depth distributions to back-project the 2D image features into 3D space, then use the pooling operation or height compression to generate BEV features.
        Others take advantage of Transformer \cite{vaswani2017attention} and use learnable object-level queries to directly predict 3D bounding boxes \cite{wang2022detr3d,liu2022petr,liu2022petrv2} or position-aware queries to produce BEV features \cite{li2022bevformer,zhou2022cross}. 
\begin{figure*}[t]
  \centering
  \includegraphics[trim=2 2 2 2, clip,width=0.76\linewidth]{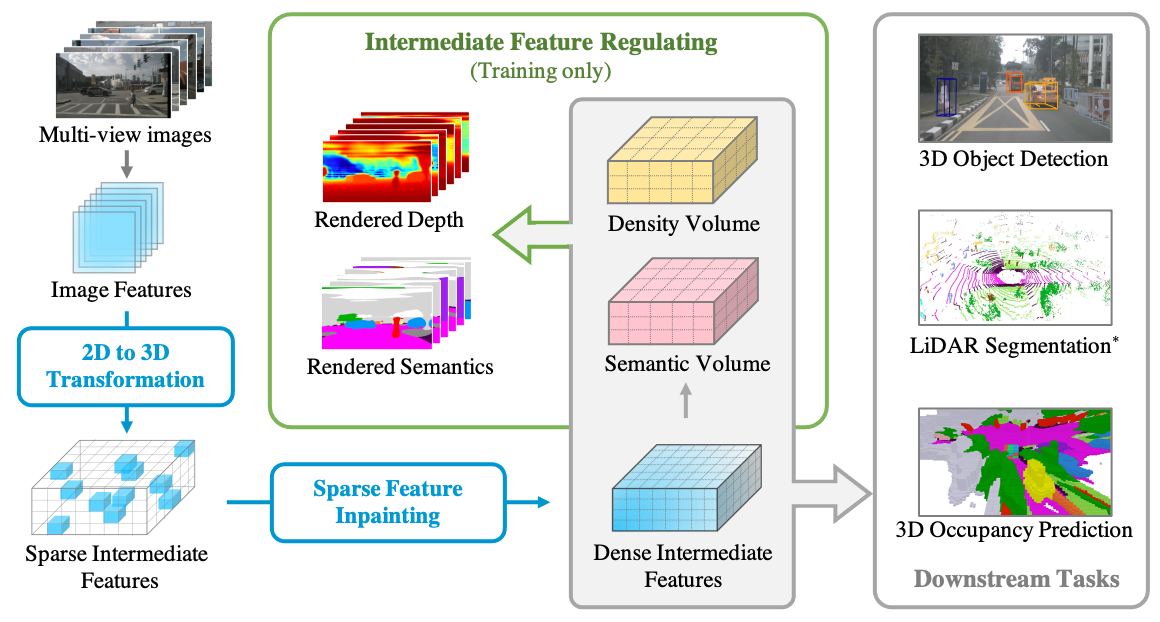}
  \caption{
\textbf{Framework of Vampire.}
We extract the 2D image features from multi-view images, 
and transform the 2D features to 3D volume space to generate sparse   intermediate 3D features.
Inpainting techniques are then applied to obtain dense intermediate features. 
Density volume and semantic volume are generated by forwarding the dense features with specific heads.
Specifically, in the training stage, we regulate intermediate features by constructing loss between ground truth and volume rendered images.
The dense features and volumes can be used for various downstream tasks such as 3D occupancy prediction, image-based LiDAR segmentation, and 3D object detection.
Please note that for the image-based LiDAR segmentation task, we do not use point clouds as input but employ its evaluation protocol~\cite{huang2023tri,wei2023surroundocc,sima2023_occnet}. 
The point clouds serve as point queries to extract features for training supervision and evaluation.
  }
  \label{fig:2}
\end{figure*}
    
        However, there are innumerable rigid and nonrigid objects with various structures and shapes in the real-world autonomous driving, which cannot be handled by classic 3D object detection.
        An alternative is to assign occupancy states to every spatial region within the perceptive range\cite{tesla2022aiday}, namely, 3D occupancy prediction.
        Unlike LiDAR segmentation \cite{fong2022panoptic} which is designed for sparse scanned LiDAR points, the occupancy prediction task aims to achieve dense 3D surrounding perception.
        This area haven't been thoroughly explored yet, only a few works use transformer-based designs to deal with it. 
        TPVFormer \cite{huang2023tri} proposes to use tri-perspective view (TPV) grid queries to interact with image features and get reasonable occupancy prediction results to describe the 3D scene. 
        SuroundOcc \cite{wei2023surroundocc} builds 3D volume queries to reserve 3D space information. CONet \cite{wang2023openoccupancy} and  SuroundOcc \cite{wei2023surroundocc} both generate dense occupancy labels for better prediction performance. 
        OccFormer \cite{zhang2023occformer} use a dual-path transformer network to get fine-grained 3D volume features. 
        Occ3D~\cite{tian2023occ3d} and OccNet~\cite{sima2023_occnet} label the original nuScenes dataset to get occupancy data at different scopes.
        In this paper, we advocate to regulate the dense 3D features to achieve better perception.
        
    \subsection{Scene Representation Learning}
        Effective 3D scene representation is the core of autonomous driving perception. 
        Voxel-based scene representations turn the 3D space into discretized voxels which is usually adopted by LiDAR segmentation \cite{ye2022lidarmultinet,ye2021drinet++,zhu2021cylindrical}, 3D scene completion \cite{cao2022monoscene,chen20203d,roldao2020lmscnet} and 3D occupancy prediction \cite{wang2023openoccupancy}. 
        BEV-based scene representations collapse 3D features onto the Bird's Eye View (BEV) plane and achieve a good balance between accuracy and efficiency. 
        They show its effectiveness in 3D object detection \cite{li2022bevformer,huang2021bevdet,li2022bevdepth,li2022bevstereo,zhang2022beverse} and BEV segmentation \cite{philion2020lift,li2022bevformer,hu2021fiery,xie2022m} but are not applicable for dense perception tasks when losing the height dimension. 
        Notably, recent implicit scene representation methods demonstrate their potential to represent meaningful 3D scenes. 
        They learn continuous functions to consume 3D coordinates and output representation of a certain point.
        This kind of representation can model scenes at arbitrary-resolution and is commonly used for 3D reconstruction \cite{chabra2020deep,park2019deepsdf} and novel view synthesis \cite{mildenhall2021nerf,barron2021mip,yariv2021volume,wang2021neus}.
        
        As far as we know, very limited researches have explored on combining implicit scene representations with existing representations in autonomous driving. 
        Tesla \cite{tesla2022aiday} firstly discusses the similarity between scene occupancy and NeRF \cite{mildenhall2021nerf} but it concentrates on using volume rendering to train the occupancy representation for scene reconstruction. 
        A recent related work is \cite{gan2023simple}, which adopts similar idea to consider the occupancy the same as volume density and use volume rendering for better depth estimation. 
        Their models use parameter-free back projection to map 2D image features to 3D volume and aggregates the 3D features with corresponding position embeddings to predict volume density and render the depth maps.
        Most related to ours is the model of ~\cite{pan2023uniocc}, which shares the same spirit and adopts rendering-based supervision as us.
        This model is very similar to Vampire, but our work goes further to demonstrate that such volume-rendering-assisted perception framework benefits multiple perception tasks and achieves competitive results with state-of-the-art approaches.

\section{Methodology}
    \subsection{Overview}
        The overall framework is illustrated in Figure \ref{fig:2}. Vampire consists of three stages: \textbf{2D-to-3D Transformation}, \textbf{Sparse Feature Inpainting} and \textbf{Intermediate Feature Regulating}. 
        The surrounding multi-camera images are first passed to 2D image backbone to extract 2D image features. 
        In 2D-to-3D transformation stage, the 2D image features are transformed from 2D image space to the 3D volume space. 
        We follow the LSS scheme \cite{philion2020lift,li2022bevdepth,li2022bevstereo,huang2021bevdet} to perform the feature mapping along depth dimension.
        To overcome the 3D feature sparsity of LSS transformation, we take a 3D hourglass net \cite{chang2018pyramid} as feature inpaintor to conduct the sparse feature inpainting and generate dense intermediate 3D features.
        The final volume features (semantic volume and density volume) can be obtained by forwarding the dense intermediate 3D features with specific heads.
        In intermediate feature regulating stage, we sample points along the ray from camera views or BEV view and get corresponding features for rendering, the rendered images and feature maps are used to construct losses to regulate the intermediate features.
        
    \subsection{2D-to-3D Transformation}
        We adopt LSS paradigm \cite{philion2020lift} to transform 2D image features to 3D features. 
        LSS-based transformations do not generate redundant features like parameter-free transformations \cite{cheng2018geometry,sitzmann2019deepvoxels,harley2019learning,harley2022simple} and are more effective than transformer-based transformations \cite{li2022bevformer,huang2023tri,wei2023surroundocc}.
        We use two simple 1-layer 2D convolution neural network (CNN) to conduct this process.
        The first one is used to predict categorical depth distribution with softmax activation, and the second one is used to lower the dimension of image features to meet our device constraints. 
        These two CNNs work together to map image features along depth axis, and we do not explicitly supervise this mapping process with depth labels.
        In this way, 2D image features are placed at the front visible surface for any certain pixels.
        
    \subsection{Sparse Feature Inpainting}
        The aforementioned 2D-to-3D transformation produce sparse intermediate 3D features, and such sparsity is not appropriate for dense prediction tasks like occupancy prediction.
        To overcome this limitation, we draw inspiration from classic image inpainting \cite{liu2018image,he2022masked} and use a 3D hourglass-like design \cite{chang2018pyramid} to inpaint the sparse intermediate 3D features $V_{sparse}$ and generate dense intermediate 3D features $V_{dense}$.
        Please refer to supplementary materials for network architecture details.

    \subsection{Intermediate Feature Regulating}
        In this stage, we use the dense intermediate 3D features $V_{dense}$ to produce two volumetrically 3D features --  density volume $V_{density}$, semantic volume $V_{semantic}$.       
        Different from \cite{mildenhall2021nerf}, we adopt the SDF (Signed Distance Function) to model the volume density $\sigma$ to facilitate the trilinear interpolation during grid sampling.
        Specifically, we predict the signed distance volume $V_{sdf}$ where each value in a position in this volume represents its distance to its nearest surface.
        Then we transform the signed distance volume $V_{sdf}$ to density volume $V_{density}$ by applying transformation function.
        We use the same transformation function as \cite{yariv2021volume}:
        
        \begin{equation}
        \begin{split}
            &V_{density} =  \alpha \Psi_\beta \left( V_{sdf} \right ), \\
            &\Psi_\beta(s) = 
                \begin{cases} 
                    \frac{1}{2} \exp ({\frac{s}{\beta}}) & \text{if } s\leq 0 \\
                    1-\frac{1}{2}\exp ({-\frac{s}{\beta}}) & \text{if } s>0
                \end{cases}
        \end{split}
        \end{equation}
        where $\alpha, \beta > 0$ are learnable parameters and $\Psi_\beta$ is the cumulative distribution function of the Laplace distribution with zero mean and $\beta$ scale, $s$ is the predicted signed distance at coordinate $\bm{x}$.
        For a coordinate $\bm{x}$ in range of interest, we can get its feature embeddings including volume density $\bm{\sigma} \left (\bm{x} \right)$ and semantic logits $\bm{s}\left (\bm{x} \right)$ by grid sampling $\mathcal{G}\left(\cdot,\bm{x}\right)$ in these 3D volume features.

        \begin{equation}\label{eq:volume-grid-sample}
            \bm{\sigma} \left (\bm{x} \right) = \mathcal{G} \left (V_{density},\bm{x} \right), \quad \bm{s}\left (\bm{x} \right) = \mathcal{G} \left (V_{semantic},\bm{x} \right)
        \end{equation}
        To compute the depth and semantics of a single pixel, we adopt similar techniques as \cite{zhi2021place, kerr2023lerf} to accumulate feature embeddings along a ray $\vec{r} = \vec{o_t} + t\vec{d}$.
        The rendering weights are calculated by: 
        \begin{equation}
        \begin{split}
            w(t) = \int_{t}& T(t)\bm{\sigma}(t)dt, \\
            &\text{where } T(t) = \exp\left (\int_{t}(-\bm{\sigma}(c))dc \right)
        \end{split}
        \end{equation}
        So the rendered feature embeddings are:
        \begin{equation}\label{volume-rendering}
            D(r) = \int_{t}w(t)r(t)dt ,\quad S(r) = \int_{t}w(t)\bm{s}(r(t))dt
        \end{equation}

        In Vampire, we conduct volume rendering in both camera view and bird's eye view.
        
        \noindent \textbf{Camera View.} 
        For camera view, we render depth and semantic maps to achieve the supervision from 2D space.
        To render a pixel, we cast a ray from the camera center through the pixel. 
        We sample $n$ depth value $\{z_i| i = 1, ...,n\}$ for a pixel $[u, v]^T$ and use known camera calibration to back-project the pixel to several 3D points $\bm{x} \in\{[x_i, y_i, z_i]^T|i = 1, ...,n\}$.
        The corresponding volume densities and semantic logits are obtained by Equation \ref{eq:volume-grid-sample}, and the depth and semantic maps can be calculated by Equation \ref{volume-rendering}.
        
\begin{figure}[h]
   \centering
    \includegraphics[trim=2 5 2 2, clip,width=0.9\linewidth]{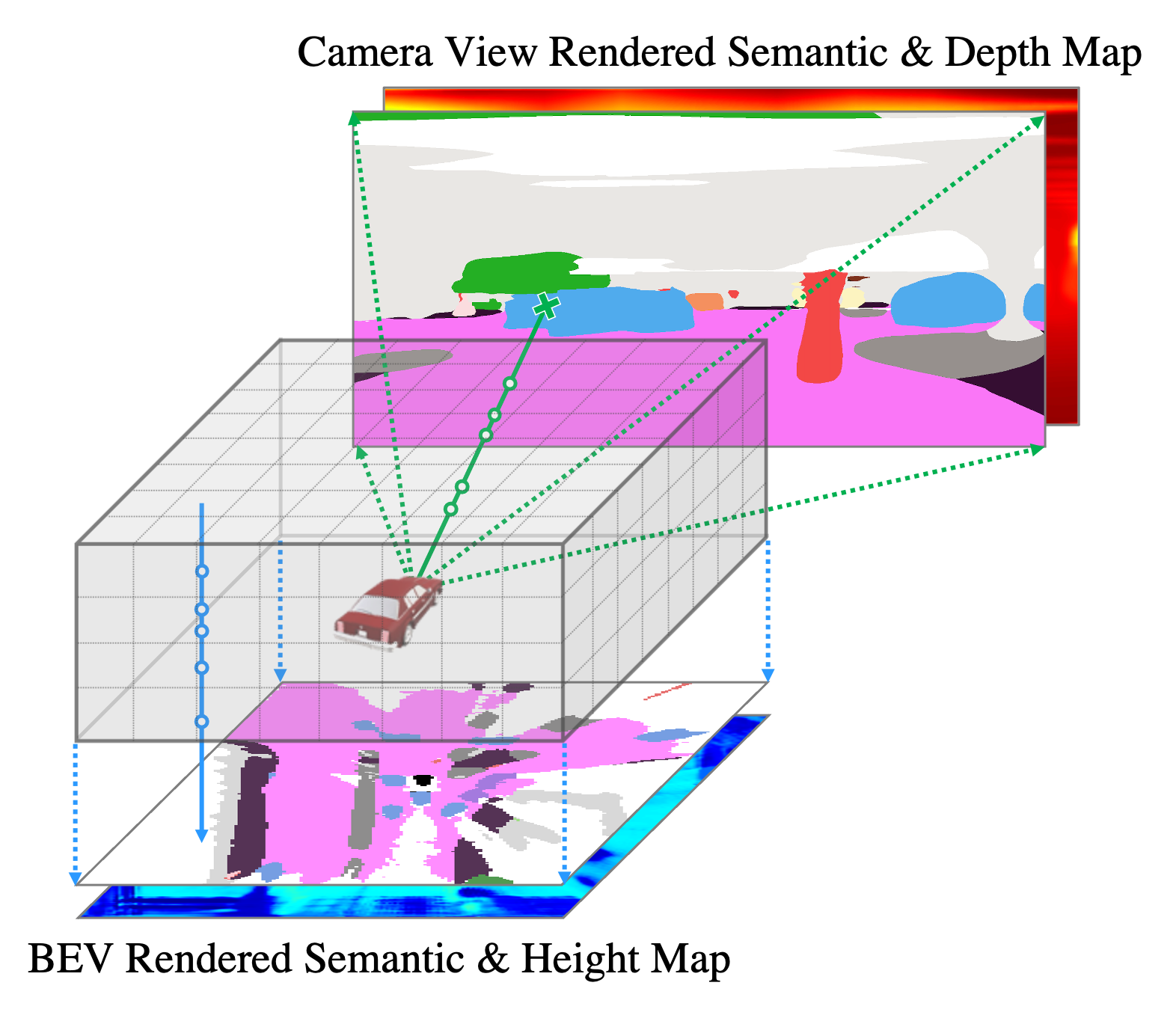}
    \vspace{-3mm}
    \caption{\textbf{Rendering operations in Intermediate Feature regulating stage.} For the camera view, the semantic and depth map are rendered by casting rays from the camera center through each pixel. 
    Several 3D points are sampled along the ray to calculate density and semantic values. For the bird's-eye-view, the semantic and height map are rendered directly from the top-down height axis.}
    \label{fig:3}
    \vspace{-5mm}
\end{figure}

        \noindent \textbf{Bird's Eye View.}
        Different from rendering in camera view, we do not need camera calibration under bird's-eye-view.
        Instead, we render directly from the top-down height axis to obtain the BEV height maps and BEV semantic maps.
        See Figure \ref{fig:3} for reference.
    \subsection{Optimization}
        \noindent \textbf{Depth Consistency Loss.} We enforce consistency between the rendered depth (or height) $D$ and the ground-truth camera depth (or BEV height) $\bar D$. 
        \begin{equation}\label{loss:depth}
        \begin{split}
            \mathcal{L}_{dep} &= \frac{1}{N_{valid}^{c}}\sum_{i=1}^{N_{valid}^{c}} {\mathrm{Smooth_{L_1}}} (D_i^c - \bar D_{i}^{c}) \\
            &+ \frac{1}{N_{valid}^{b}}\sum_{i=1}^{N_{valid}^{b}} {\mathrm{Smooth_{L_1}}} (D_i^b - \bar D_{i}^{b}) 
        \end{split}
        \end{equation}
        \noindent \textbf{Semantic Consistency Loss.} Similarly, we impose consistency on the volume-rendered semantic logits $S^{c}$ and the ground-truth semantic label $\bar S^{c}$, we employ both cross entropy (CE) loss and lovasz-softmax (LS) loss:
         \begin{equation}\label{loss:semantic}
         \begin{split}
            \mathcal{L}_{sem} &=  \frac{1}{N_{valid}^{c}}\sum_{i=1}^{N_{valid}^{c}} (\text{CE} (S_i^{c}, \bar S_{i}^{c}) + \text{LS} (S_i^{c}, \bar S_{i}^{c})) \\
            &+  \frac{1}{N_{valid}^{b}}\sum_{i=1}^{N_{valid}^{b}} (\text{CE} (S_i^b, \bar S_{i}^{b}) + \text{LS} (S_i^b, \bar S_{i}^{b}))   
        \end{split}
        \end{equation}
        where $N_{valid}^{c}$ is the number of pixels with ground-truth depth for all cameras (obtained by projecting the sparse LiDAR points to current camera image plane). $N_{valid}^{b}$ is the number of pixels with ground-truth height in the range of BEV. 
        $\bar D^{c}$ and $\bar S^{c}$ are the ground-truth depths and semantic labels obtained by projecting the sparse LiDAR points to the current camera image plane.
        For BEV, we get the ground-truth label $\bar D^b$ and $\bar S^{b}$ by projecting the LiDAR points to the grounding plane and take the height and semantic label of the highest point for a pixel in BEV map.

        The overall loss we use to regulate our intermediate 3D features is:
        \begin{equation}
            \mathcal{L}_{reg} = \lambda_{dep}\mathcal{L}_{dep} + \lambda_{sem}\mathcal{L}_{sem}
        \end{equation}
        where $\lambda_{dep}, \lambda_{sem}$ are fixed loss weights. We empirically set all weights to 1 by default.
        
    \subsection{Applications of Vampire Features}
        We follow the existing scheme \cite{huang2023tri,li2022bevdepth} to use our regulated intermediate features. 
        
        \noindent \textbf{3D occupancy prediction.} The 3D occupancy prediction task usually covers a certain range of scene, thus we voxelize the interested scene range and conduct grid sampling from our predicted semantic volume $V_{semantic}$ with these voxel center coordinates. We use the output semantic logits to represent the semantic occupancy for each voxel.

        \noindent \textbf{LiDAR segmentation.} Different from traditional LiDAR segmentation task, our model consumes purely RGB images to perceive 3D surroundings rather than LiDAR point cloud. 
        To conduct LiDAR segmentation, we use LiDAR point clouds as point queries to get the corresponding semantic logits from semantic volume $V_{semantic}$.
        
        \noindent \textbf{3D object detection.} We adopt a tanh function to scale the density volume $V_{density}$ to the range of [0, 1] and then use it to enhance dense 3D intermediate features $V_{dense}$.
        We collapse the height dimension and use a linear layer to squeeze the feature dimension:
        \begin{equation}
            F_{BEV} = \mathcal{HC}(V_{dense} \cdot \tanh(V_{density}))
        \end{equation}
        Where $\mathcal{HC}$ stands for ``height compression'', which collapses the height dimension of 3D features then squeezes the feature dimension with a linear layer to get the final BEV shape features $F_{BEV}$ for detection.
        $F_{BEV}$ is then fed into the detection head to obtain final detection results.
        For simplicity, we adopt the detection head of BEVDepth \cite{li2022bevdepth} to produce 3D object detection results.
          
\section{Experiments}
    To evaluate the proposed method, we benchmark Vampire on challenging public autonomous driving datasets nuScenes \cite{caesar2020nuscenes} and its variants \cite{tian2023occ3d,fong2022panoptic}.
    
        \noindent \textbf{Datasets.} 
        The nuScenes dataset contains 1000 scenes of 20 seconds duration each, and the key samples are annotated at 2Hz. 
        Each sample consists of RGB images from 6 surrounding cameras with 360° horizontal FOV and point cloud data from 32 beams LiDAR.
        The total of 1000 scenes are officially divided into training, validation and test splits with 700, 150 and 150 scenes, respectively.
        Occ3D-nuScenes~\cite{tian2023occ3d} contains 700 traing scenes and 150 validation scenes. The occupancy scope is defined as $[-1.0, 5.4] \times [-40.0, 40.0] \times [-40.0, 40.0] (meter)$ with a voxel size of 0.4-meter.
        

        \noindent \textbf{Implementation details.} 
        Our implementation is based on official repository of BEVDepth~\cite{li2022bevdepth}. 
        We use ResNet-50 \cite{he2016deep} as image backbone and the image resolution of 256 $\times$ 704 to meet our computational resources.
        For the inpainting network, we adpot an hourglass-like architecture (further details are provided in our supplementary materials).
        The intermediate 3D feature resolutions are $20\times256\times256$ corresponding to the range of $[-3.0, 5.0] \times [-51.2, 51.2] \times [-51.2, 51.2] (meter)$ and the 3D feature dimension are set to 16 by default.
        We use AdamW as an optimizer with a learning rate set to 2e-4 and weight decay as 1e-7. 
        All models are trained for 24 epochs with a total batch size of 8 on 8 3080Ti GPUs (12GB).
\begin{table}[h]
	\setlength\tabcolsep{5pt}
	\centering
	\begin{tabular}{l|c|c|c}
		\toprule
		Method
		& Backbone & \makecell{Image \\ Size} & mIoU $\uparrow$
		\\
		\midrule
        MonoScene~\shortcite{tian2023occ3d} & Effi.NetB7 & {900$\times$1600} & 6.1 \\
        \midrule
        BEVDet~\shortcite{tian2023occ3d} & \multirow{5}{*}{R101}& \multirow{5}{*}{900$\times$1600} &19.4 \\
        OccFormer~\shortcite{tian2023occ3d} &  &  &21.9 \\
        BEVFormer~\shortcite{tian2023occ3d} &  &  & 26.9  \\
        TPVFormer~\shortcite{tian2023occ3d} &  &  &27.8 \\
        CTF-Occ~\shortcite{tian2023occ3d} &  &  & \textbf{28.5} \\
        \midrule
        UniOcc~\shortcite{pan2023uniocc} & \multirow{2}{*}{R50} & \multirow{2}{*}{256$\times$704} & 22.0 \\
        \bf Vampire (ours) & & & \underline{28.3} \\
		\bottomrule
	\end{tabular}
    \vspace{-1.5mm}
     \caption{\textbf{3D occupancy prediction results on Occ3D-nuScenes.} 
     ``Effi.NetB7'' stands for EfficientNetB7.
     We obtain the values of other methods from the benchmark paper~\cite{tian2023occ3d}. 
     We use \textbf{bold} to indicate the highest result and \underline{underline} for the second-best result.
     Despite image backbone and input size differences, Vampire achieves comparable performance with state-of-the-art methods.}
	\label{tab:occ-simple}
 
\end{table}        
    \subsection{3D Occupancy Prediction}

        We compare Vampire with previous state-of-the-art methods on the 3D occupancy prediction task in Table \ref{tab:occ-simple}. 
        These baseline methods including two main-stream BEV models $-$ BEVDet~\cite{huang2021bevdet}, BEVFormer~\cite{li2022bevformer} and five existing 3D occupancy prediction methods $-$ MonoScene~\cite{cao2022monoscene}, TPVFormer~\cite{huang2023tri}, OccFormer~\cite{zhang2023occformer}, UniOcc~\cite{pan2023uniocc}, and CTF-Occ~\cite{tian2023occ3d}. 
        It can be observed that our method achieves comparable performance with these methods under the mIoU metric.
        Our Vampire surpasses OccFormer / BEVFormer / TPVFormer by 6.4 / 1.4 / 0.5 mIoU. 
        Although Vampire has a lower mIoU than CTF-Occ (28.3 v.s. 28.5), it is still promising since our method adopts a relatively weak image backbone ResNet-50 and lower input image resolution (256 $\times$ 704). 

\vspace{-3mm}
\begin{table}[h]
	\setlength\tabcolsep{5pt}
	\centering
	\begin{tabular}{l|c|c|c}
		\toprule
		Method
		& Backbone & \makecell{Image \\ Size} & mIoU $\uparrow$
		\\
        \midrule
        BEVFormer~\shortcite{huang2023tri} &  \multirow{4}{*}{R101} & \multirow{4}{*}{900$\times$1600} & 56.2  \\
        TPVFormer~\shortcite{huang2023tri} &  &  & \textbf{68.9} \\
        TPVFormer\textsuperscript{$\dagger$}~\shortcite{sima2023_occnet} &  &  & 58.5 \\
        OccNet\textsuperscript{$\dagger$}~\shortcite{sima2023_occnet} &  &  & 60.5 \\
        \midrule
        TPVFormer~\shortcite{huang2023tri} &  \multirow{4}{*}{R50} & 450$\times$800 & 59.3 \\
        OccNet\textsuperscript{$\dagger$}~\shortcite{sima2023_occnet}  & & 900$\times$1600 & 53.0 \\
        \bf Vampire (ours) &  & {256$\times$704} & \underline{66.4} \\
        \bf Vampire (ours) \textsuperscript{$\dagger$} &  & {256$\times$704} &62.2 \\
		\bottomrule
	\end{tabular}
    \vspace{-1.5mm}
     \caption{\textbf{LiDAR segmentation results on Panoptic nuScenes~\cite{fong2022panoptic} validation set.} 
     We obtain the values of baselines from their respective papers. 
     We use \textbf{bold} to indicate the highest result and \underline{underline} for the second-best result.
     Mark $\dagger$ indicates methods trained without direct LiDAR supervision but only occupancy semantic labels. 
     }
     \vspace{-5mm}
	\label{tab:lidar-seg-simple}
\end{table}
    \subsection{LiDAR Segmentation}
        We compare Vampire with existing image-based LiDAR segmentation methods in Table \ref{tab:lidar-seg-simple}.
        These baseline methods including BEVFormer~\cite{li2022bevformer}, TPVFormer~\cite{huang2023tri} and OccNet~\cite{sima2023_occnet}.
        In the inference stage, we predict the semantic labels for given points in the LiDAR segmentation task. 
        Vampire surpasses the SOTA model TPVFormer~\cite{huang2023tri} with the same backbone in terms of mIoU (66.4 v.s. 59.3), but a little lower (-2.5) compared to TPVFormer-R101.
        Even without direct 3D LiDAR supervision, Vampire can outperform OccNet~\cite{sima2023_occnet} models with different backbones respectively by 9.2 and 1.7 points in mIoU.
        
\begin{table}[ht]
	\setlength\tabcolsep{4pt}
	\centering
	\begin{tabular}{l|c|c|c|c}
		\toprule
		Method  & Joint. & mAP $\uparrow$ & NDS $\uparrow$ & mAVE $\downarrow$
		\\
        \midrule
        BEVFormer~\shortcite{li2022bevformer}  &  & 0.257 & 0.359  &0.660\\
        BEVDet~\shortcite{li2022bevdepth}   & & 0.286 &0.372 & - \\
        BEVDetph~\shortcite{li2022bevdepth}   &  & 0.322 &0.367 & -\\
        \midrule
        BEVNet~\shortcite{sima2023_occnet}  & \multirow{4}{*}{\checkmark} &0.271 &\textbf{0.390} &\bf 0.541\\
        VoxNet~\shortcite{sima2023_occnet}    &  &0.277 & 0.387 &0.614\\
        OccNet~\shortcite{sima2023_occnet}    & &0.276 & \textbf{0.390} &0.570\\
        \bf Vampire (ours)   & & \textbf{0.301} & 0.354 &1.043\\
		\bottomrule
	\end{tabular}
    \vspace{-1.5mm}
     \caption{\textbf{3D object detection results on nuScenes validation set.}
     ``mAVE'' stands for mean Average Velocity Error.
     Vampire achieves comparable mAP with baseline methods but fails to sense accurate velocity. 
     The joint-training baselines are trained with additional occupancy flow annotation (occupancy velocity)~\cite{sima2023_occnet}, which can significantly improve their performance to perceive object speed.
     }
	\label{tab:det-simple}
    \vspace{-5mm}
\end{table} 
    \subsection{3D Object Detection}
        We conduct 3D object detection experiments on nuScenes validation set.
        The intention is to verify whether the regulated 3D features can still qualified for 3D detection task. 
        We choose several main-stream 3D object detection baselines including BEVFormer~\cite{li2022bevformer}, BEVDet~\cite{huang2021bevdet} and BEVDepth \cite{li2022bevdepth}.
        For fair comparisons, we report the baseline values under the setting of ResNet-50 backbone and without temporal fusion techniques.
        We also choose three baselines provided by~\cite{sima2023_occnet} which conducts joint training of occupancy prediction and 3D detection task like us. 
        As shown in Table~\ref{tab:det-simple}, comparing to normal 3D object detection methods, Vampire surpasses BEVFormer and BEVDet in mAP (0.301 v.s. 0.286), but lower in NDS (0.354 v.s. 0.372).
        This could be attributed to negative transfer~\cite{pan2009survey} in joint training of multi-task.
        BEVDepth reports the value with EMA technique and a large batch size of 64, thus we attribute the performance gap to that. 
        For joint training baselines, Vampire achieves a significantly higher mAP (0.301 v.s. 0.277), but has a gap on the metric of NDS (0.354 v.s. 0.390) and the metric of mean Average Velocity Error (0.541 v.s. 1.043).
        To summarize, Vampire can perceive the geometry details of 3D surroundings but less sensitive with object velocity. 
        It is because the baseline methods are trained by occupancy data with additional flow annotation (occupancy velocity), which can significantly improve their performance to perceive object speed.

\begin{table}[h]
        \vspace{-3mm}
	\setlength\tabcolsep{5pt}
	\centering
	\begin{tabular}{c c  c c | c c c}
		\toprule
	    Trans. & Inp. & $\mathcal{L}_{dep}$ &$\mathcal{L}_{sem}$ & Occ.$\uparrow$ & Seg.$\uparrow$ & Det.$\uparrow$ \\
		\midrule
        Bilinear & & & & 21.3 & 56.7 & 0.301\\ 
        LSS & & & & 21.9 & 56.5 & \textbf{0.318} \\
        LSS & \checkmark & & & 23.8 & 60.1 & 0.316 \\
        LSS & \checkmark & \checkmark & &  24.9 & 59.6 & 0.309\\
        LSS & \checkmark & \checkmark & \checkmark & \textbf{25.8} &\textbf{62.6} &\textbf{0.318}\\
		\bottomrule
	\end{tabular}
    \vspace{-1.5mm}
 \caption{\textbf{Ablation study for network structures and losses.}
 ``Trans.'' stands for 2D-to-3D transformation.
 ``Inp.'' stands for sparse feature inpainting.
  ``Occ.'' represents 3D occupancy prediction, ``Seg.'' refers to LiDAR segmentation, ``Det.'' denotes 3D object detection.
	}
 \vspace{-5mm}
	\label{tab:ablation}
\end{table}
    \subsection{Ablation Studies}
        
        \noindent
        \textbf{Architectural components.} We conduct an ablation study on network structures and the proposed losses under the multi-task setting in Table~\ref{tab:ablation}. 
        For 3D occupancy prediction task and LiDAR segmentation task, we report the mIoU.
        For 3D object detection task, we report the NDS.
        As a parameter-free method, Bilinear~\cite{harley2022simple} can produce dense 3D features in the simplest way but also cause massive features generated at the wrong 3D spaces, resulting poor performances in all three tasks.
        The LSS baseline produces sparse 3D intermediate features, which can handle object detection, but fails to handle dense prediction tasks (\textit{e.g.}, occupancy prediction).
        When employing the feature inpaintor, dense point / grid level tasks (\textit{i.e.}, occupancy and segmentation) obtain significant improvements.
        The regulation of depth $\mathcal{L}_{dep}$ improves the occupancy prediction, but has a negative effect for LiDAR segmentation and detection.
        Such negative effect is because $\mathcal{L}_{dep}$ imposes constraints to the density volume $V_{density}$ and enhances both foreground (\textit{e.g.}, cars) and background objects (\textit{e.g.}, trees).
        $\mathcal{L}_{sem}$ provides extra semantic information, which alleviates the performance drops and achieve the best results.
        
\begin{table}[h]
\vspace{-3mm}
	\centering
	\begin{tabular}{c c| c c c}
		\toprule
	    Camera. & BEV. & Occ.$\uparrow$ & Seg.$\uparrow$ & Det.$\uparrow$ \\
		\midrule
        \checkmark &  & 24.6 & 61.9 & 0.303\\ 
        & \checkmark & 24.9 & 60.0 & 0.315 \\
        \checkmark & \checkmark & \bf 25.8 &\bf 62.6 &\bf 0.318\\
		\bottomrule
	\end{tabular}
 \vspace{-1.5mm}
 \caption{\textbf{Ablation study for camera and BEV views.}
 ``Camera.'' stands for volume rendering loss in camera view.
 ``BEV.'' stands for volume rendering loss in BEV view.
  ``Occ.'' represents 3D occupancy prediction, ``Seg.'' refers to LiDAR segmentation, ``Det.'' denotes 3D object detection.
	}
	\label{tab:ablation-view}
\vspace{-2mm}
\end{table}
        \noindent
        \textbf{Supervision of different views.} We provide the ablation experiments for both views. 
        As shown in Table~\ref{tab:ablation-view}, LiDAR segmentation is more relevant with the supervision from camera view and 3D object detection is more sensitive to the supervision of BEV view.
        The camera view supervision can provide fine-grained geometry information which facilitates the LiDAR segmentation.
        However, the upper parts of camera view has very few LiDAR points for supervision (no LiDAR in the sky), thus the upper parts of density and semantic volumes are out of control. 
        This could explain the degradation of detection performance when only supervising the camera view. 
        BEV view can provide extra information and squeezing the upper parts of $V_{density} / V_{semantic}$ to meet their highest surface, such occlusion information is invisible in camera views and can restrain the degradation of detection.

\begin{table}[h]
\vspace{-3mm}
	\setlength \tabcolsep{0.5pt}
	\centering
	\begin{tabular}{l | c | c c c}
		\toprule
	    Method & Device & Params. $\downarrow$& Memory $\downarrow$ &FPS$\uparrow$\\
		\midrule
        BEVNet~\shortcite{sima2023_occnet} & \multirow{3}{*}{V100}& 39M & 8G & 4.5\\ 
        VoxNet~\shortcite{sima2023_occnet} & & 72M & 23G& 1.9\\
        OccNet~\shortcite{sima2023_occnet} & & 40M & 18G& 2.6\\
        \midrule
        BEVFormer~\shortcite{wei2023surroundocc} & \multirow{3}{*}{RTX3090}& - & 4.5G & 3.2\\
        TPVFormer~\shortcite{wei2023surroundocc} & & - & 5.1G & 3.1 \\
        OccFormer~\shortcite{zhang2023occformer} & & 147M & 5.9G & 2.9 \\
        \midrule
        \bf Vampire (ours) & RTX3080Ti & 52M &5.0G &3.8 \\
		\bottomrule
	\end{tabular}
 \vspace{-1.5mm}
 \caption{\textbf{Efficiency analysis.} 
 The experiments are all conducted with the corresponding device.
	}
	\label{tab:efficiency}
 \vspace{-5mm}
\end{table}
    \subsection{Efficiency Analysis.}   

        In Table~\ref{tab:efficiency}, we compare the inference latency and memory of several methods. 
        Due to our hardware constraints, we do not run these methods on our own but crop the reported values from several papers.
        We obtain the values from~\cite{sima2023_occnet} for BEVNet, VoxNet and OccNet.
        We obtain the values from~\cite{wei2023surroundocc} for BEVFormer, TPVFormer and OccFormer.
        We find that the computational resources used in our methods are moderate.
        This makes the method practical and easy to use for the community.
                
    \subsection{Qualitative Results}
\begin{figure}[h]
    \centering
    \begin{subfigure}{\linewidth}
        \centering
        \includegraphics[width=0.9\linewidth]{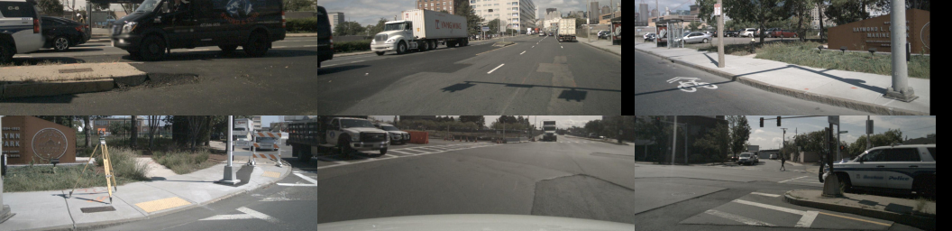}
        \caption{Input multi-view images.}
        \label{fig:4-a}
    \end{subfigure}
    \begin{subfigure}{\linewidth}
        \begin{minipage}{\linewidth}
        \centering
        \includegraphics[height=15mm]{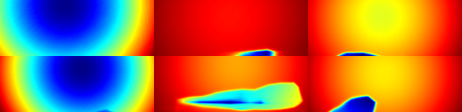}
        \includegraphics[height=15mm]{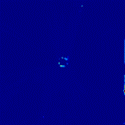}
        \end{minipage}
        \begin{minipage}{\linewidth}
        \centering
        \includegraphics[height=15mm]{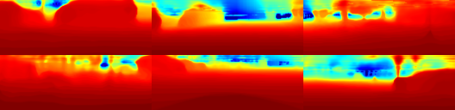}
        \includegraphics[height=15mm]{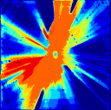}
        \end{minipage}
        \caption{Rendered depth / density maps without/with $\mathcal{L}_{dep}$.}
        \label{fig:4-b}
    \end{subfigure}
    \begin{subfigure}{\linewidth}
        \begin{minipage}{\linewidth}
        \centering
        \includegraphics[height=15mm]{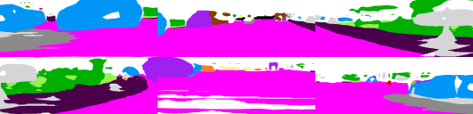}
        \includegraphics[height=15mm]{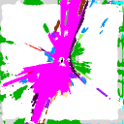}
        \end{minipage}
        \begin{minipage}{\linewidth}
        \centering
        \includegraphics[height=15mm]{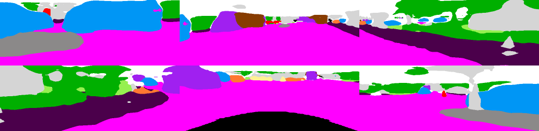}
        \includegraphics[height=15mm]{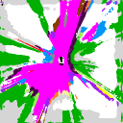}
        \end{minipage}
        \caption{Rendered camera / BEV semantic maps without/with $\mathcal{L}_{sem}$.}
        \label{fig:4-c}
    \end{subfigure}
    \vspace{-5.5mm}
    \caption{\textbf{Visualizations of rendered results.}
    The effectiveness of $\mathcal{L}_{dep}$ can be verified by Figure \ref{fig:4-b}, $\mathcal{L}_{dep}$ imposes constraints for learning reasonable 3D geometry information.
    The effectiveness of $\mathcal{L}_{sem}$ can be verified by Figure \ref{fig:4-c}, semantic regulation provides significant improvements in generating dense and meaningful features.
    It is obvious that regulations from depth and semantic offer incremental gains for these qualitative results.}
    \label{fig:4}
    \vspace{-5mm}
\end{figure}

        In Figure \ref{fig:4}, we provide visualizations of the proposed regulations from depth and semantics. 
        Due to the space constraints, we provide additional qualitative results in the supplementary. 
        3D occupancy results are best viewed as videos, so we urge readers to view our supplementary videos.
\section{Conclusion}
    In this paper, we explore the connections between space occupancy in autonomous driving and volume density in NeRF, and propose a novel vision-centric perception framework, \textit{i.e.}, Vampire, which takes volume rendering as the intermediate 3D feature regulator in the multi-camera setting. 
    Vampire predicts per-position occupancy as the volume density and accumulate the intermediate 3D features to 2D planes to obtain additional 2D supervisions.
    Extensive experiments show that our method is competitive with existing state-of-the-arts across multiple downstream tasks.
    
\section{Acknowledgement}
    \noindent This work was supported in part by The National Nature Science Foundation of China (Grant Nos: 62303406, 62273302, 62036009, 61936006), in part by Ningbo Key R\&D Program (No.2023Z231, 2023Z229), in part by Yongjiang Talent Introduction Programme (Grant No: 2023A-194-G), in part by the Key R\&D Program of Zhejiang Province, China (2023C01135).
\bibliography{aaai24}

\clearpage
\appendix
\begin{center}
	\textbf{\Large Appendix}
\end{center}

\renewcommand\thesection{\Alph{section}}

Considering the space limitation of the main text, we provide additional results and discussion in this supplementary material, which is organized as follows:
\begin{itemize}
    \item Section {\ref{sec:disscussion}}: Explanations and Discussions
    \begin{itemize}
        \item Implementation details. \ref{ssec:implementation}
        \item Discussions. \ref{ssec:discussion}
        \item Limitations and future work. \ref{ssec:limitation}
        \item Board impact. \ref{ssec:board}
        \item Code. \ref{ssec:code}
    \end{itemize}
    
    \item Section {\ref{sec:additional}}: Additional Quantitative Results
    \begin{itemize}
        \item Detailed results of 3D occupancy prediction. \ref{ssec:occ-full}
        \item Detailed results of LiDAR segmentation. \ref{ssec:lidar-seg-full}
        \item Detailed results of 3D object detection. \ref{ssec:det-full}
        \item Ablation of position information. \ref{ssec:ablation-pos}
        \item Training resource comparison. \ref{ssec:train-resouce}
    \end{itemize}
    
    \item Section {\ref{sec:qualitative}}: Additional Qualitative Results
    \begin{itemize}
        \item Detection results on nuScenes. \ref{ssec:qualitative-nuscenes}
        \item Rendered results and 3D visualizations. \ref{ssec:rendered-results}
        \item Occupancy demo video. \ref{ssec:occupancy-results}
    \end{itemize}
    
\end{itemize}
\section{Explanations and Discussions}\label{sec:disscussion}
    \subsection{Implementation details.}\label{ssec:implementation}
        Vampire adopts ResNet-50 \cite{he2016deep} pretrained on ImageNet~\cite{deng2009imagenet}.
        The 2D image features are transformed into 2.5D frustum features using a 2D convolutional neural network (CNN) layer with softmax activation. 
        The depth channel dimension of this CNN layer is 86, covering a range from 2.0 to 70.4 meters.  This range is significantly longer than that of the intermediate 3D features, as not all 2D image features should be placed in the 3D feature range.
        Then we use pre-defined voxel coordinates ($[-3.0, 5.0] \times [-51.2, 51.2] \times [-51.2, 51.2] (meter)$) to grid-sample~\cite{reading2021categorical} these frustum features to obtain sparse 3D features.
        Before inpainting the sparse 3D features, they are concatenated with their normalized 3D coordinates in the range of [-1, 1] to incorporate position information. 
        This addition of 3D position information has been proven effective in previous works such as ~\cite{liu2022petr,liu2022petrv2,zhou2022cross}, and we also provide experiments to validate its effectiveness in Section \ref{ssec:ablation-pos}.
        We use an hourglass-like network~\cite{chang2018pyramid} to perform sparse feature inpainting to obtain dense 3D features. 
        
\begin{figure}[!t]
  \centering
  \vspace{-3mm}
  \includegraphics[trim=4 4 4 4, clip,width=\linewidth]{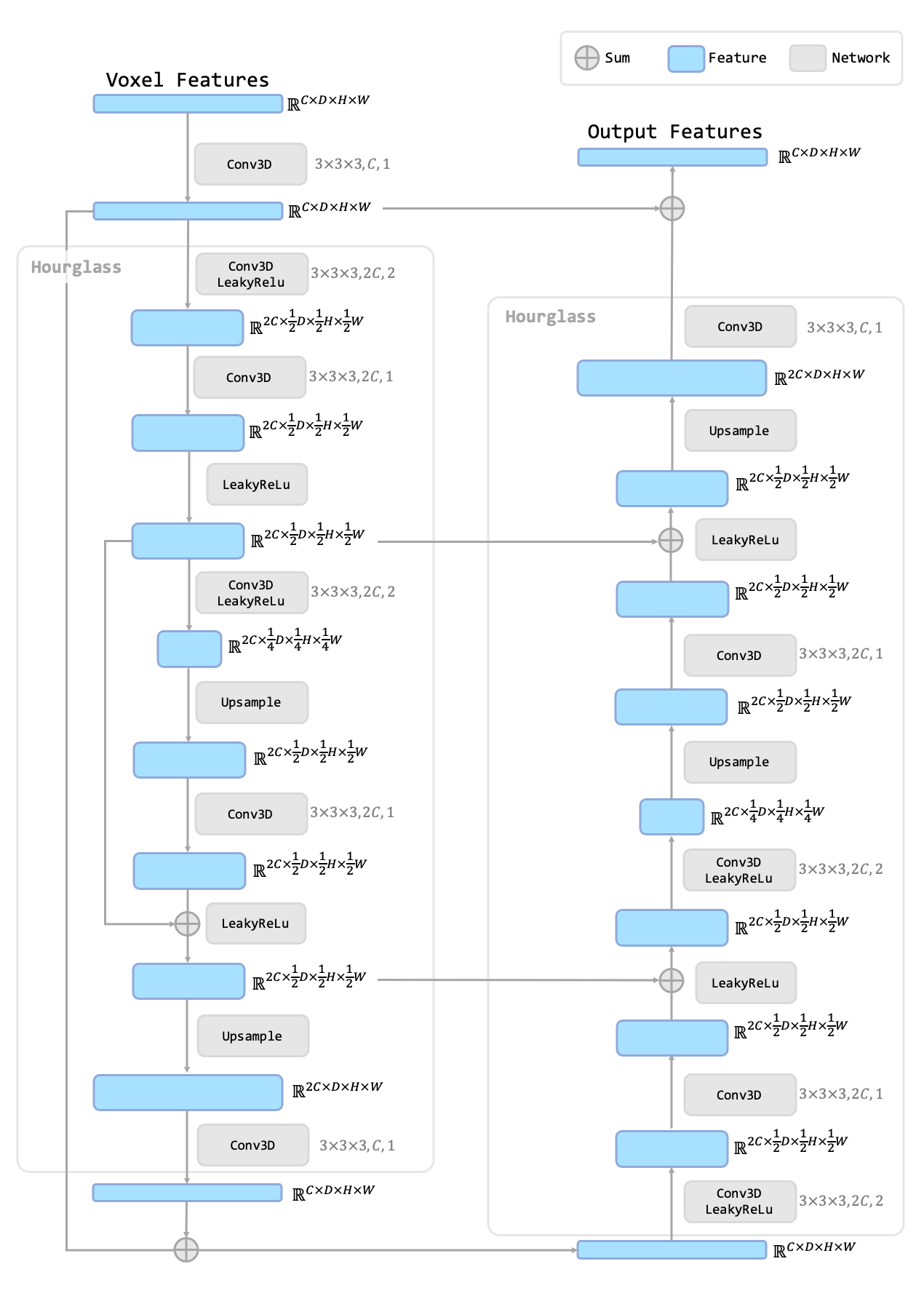}
  \caption{\textbf{Sparse feature inpaintor.} We adpot a 3D hourglass-like network to conduct feature inpainting.
  }
  \vspace{-5mm}
  \label{supply-fig:1}
\end{figure}

        The architecture of the inpaintor network is illustrated in Figure \ref{supply-fig:1}.
        Then we employ two one-layer 3D convolutions with kernel size 3  to generate the SDF volume $V_{sdf}$ and semantic volume $V_{semantic}$.
        Subsequently, the SDF volume is processed using the transformation function described in Equation 1 in the main text. 
        This process yields the density volume, denoted as $V_{density}$.
        During the rendering process, we utilize the $1/4$ downsampled features, specifically depth maps and semantic maps, rather than the original resolution. This downscaling is implemented due to GPU memory constraints.

    \subsection{Discussions.}\label{ssec:discussion}
        \noindent
        {\bf \textit{@ Why not Render RGB Images for Supervision?}}

        \noindent
        We abstain from predicting an RGB volume and rendering RGB images for several reasons. First, RGB images in autonomous driving datasets are often captured under varying illumination conditions, leading to inconsistent RGB information compared to the more stable depth and semantic information across day and night scenes.
        Second, RGB color strongly correlates with the viewing direction. 
        In the case of NeRF, RGB color is encoded with respect to both spatial position and viewing direction. 
        However, in vision-based autonomous driving scenarios, camera viewing directions tend to remain relatively constant, which decouples RGB color and view directions. 
        This decoupling of RGB color and viewing directions simplifies the modeling task.
        Third, due to limitations in GPU memory, our intermediate 3D features are constrained to a feature dimension of 16. 
        Incorporating RGB information into these intermediate 3D features, alongside volume density and semantics, becomes challenging within these memory constraints.\\
        \\
        \noindent    
        {\bf \textit{@ Why Adopt SDF-Based Density Modeling?}}
      
        \noindent
        In the process of obtaining the 3D feature for a specific position, we employ the grid-sampling operation within the corresponding feature volume. 
        If we were to model density (occupancy) simply as a binary state (occupied or not), it would introduce substantial quantization errors. 
        The adoption of SDF-based density modeling enables us to mitigate these errors and facilitate trilinear interpolation during grid sampling.\\
        \\
        \noindent
        {\bf \textit{@ Why is there a Discrepancy in Occupancy Prediction Results Compared to the CVPR2023 3D Occupancy Challenge Benchmark~\footnote{\url{https://github.com/CVPR2023-3D-Occupancy-Prediction/CVPR2023-3D-Occupancy-Prediction}}?}}

        \noindent
        We utilized the same occupancy dataset, namely the Occ3d-nuScenes~\cite{tian2023occ3d}, as the one employed in the CVPR2023 3D Occupancy Challenge. 
        The Occ3d-nuScenes dataset provides binary camera visibility masks denoted as $[\mathrm{mask\_camera}]$, indicating whether voxels are observable within the current camera viewpoints. 
        The inclusion of these masks during training significantly aids the learning process but may potentially lead to reduced visualization quality~\cite{pan2023uniocc}. 
        This trade-off between prediction performance and visualization quality arises because the model may tend to neglect areas beyond the visible region.

        There have been ongoing discussions concerning the use of such visibility masks\footnote{\url{https://github.com/Tsinghua-MARS-Lab/Occ3D/issues/3}}\footnote{ \url{https://github.com/CVPR2023-3D-Occupancy-Prediction/CVPR2023-3D-Occupancy-Prediction/issues/40}}. 
        Given that the competition evaluation exclusively considers visible voxels, most of the methods participating in the benchmark have adopted these visibility masks for training. To align with the official paper on the Occ3d-nuScenes dataset~\cite{tian2023occ3d}, we conducted experiments without the use of $[\mathrm{mask\_camera}]$.\\
        \\
        

        \noindent
        {\bf \textit{@ Why Do Joint-Training Methods Fail to Exhibit Significant Advantages Compared to Conventional 3D Object Detection Methods?}}
       
        \noindent
        Contemporary state-of-the-art 3D object detection methods often leverage high-resolution input images, robust backbone architectures, and an array of intricate engineering techniques, data augmentations, Exponential Moving Averages (EMA), CBGS, multi-frame fusion (sometimes spanning into future frames), test-time augmentations, and ensemble strategies. 
        The attainment of impressive benchmark results with these methods is prevalent. 
        However, it's worth noting that many of these techniques necessitate substantial additional GPU memory resources, which can be beyond the constraints of available hardware. 
        For instance, our computational resources are limited to 8 RTX 3080Ti GPU.
        
        As described in the main text, we attribute the performance gap between joint-training methods and conventional ones to the EMA technique and the use of large batch sizes. 
        To emphasize the advantages of our proposed method, we establish a fair baseline on the official BEVDepth \href{https://github.com/Megvii-BaseDetection/BEVDepth}{repository}. This baseline omits the use of EMA and multi-frame fusion techniques, employing a total batch size identical to ours (i.e., 8). 
        In Table \ref{tab:vampire-vs-bevdepth}, we present the experimental results, showcasing the superior performance of our proposed approach when compared to the baseline method used in BEVDepth.
        
\begin{table}[ht]
    \scriptsize
	\setlength\tabcolsep{1pt}
	\centering
	\begin{tabular}{l|c c| c c c c c}
		\toprule
	    Method  & mAP$\uparrow$ & NDS $\uparrow$& mATE $\downarrow$ & mASE $\downarrow$ &mAOE $\downarrow$ & mAVE $\downarrow$ & mAAE $\downarrow$\\
        \midrule
        BEVDepth\textsuperscript{$*$} \shortcite{li2022bevdepth} &\bf{0.244} & 0.308 & 0.769 & 0.300 & 0.787 & 1.045 & 0.280 \\
        \bf{Vampire} (ours)\textsuperscript{$*$} & 0.210 & \bf{0.318} & 0.739 & \bf{0.291} & 0.669 & \bf{0.922} &\bf{0.253} \\
        \midrule
        BEVDepth \shortcite{li2022bevdepth} &0.291 & 0.343 & 0.745 & \bf 0.277 & \bf 0.646 & 1.163 & 0.354 \\
        \bf{Vampire} (ours) & \bf 0.301 & \bf 0.354 &\bf 0.741 &0.278 &0.649 &\bf 1.043 &\bf 0.297 \\
		\bottomrule
	\end{tabular}
    \caption{
      \textbf{Comparison between Vampire and BEVDepth.}
      We have established a baseline for the sake of convenience and fair comparison. 
      It is noteworthy that all models presented in this table were trained using a total batch size of 8 to ensure that the results are not skewed by the influence of larger batch sizes. Additionally, the asterisk (*) next to a method denotes that it was trained without the application of data augmentation techniques. 
      Our method consistently outperforms BEVDepth, both with and without the use of data augmentations.
    	}
	\label{tab:vampire-vs-bevdepth}
\end{table}
  
    \subsection{Limitations and future works.}\label{ssec:limitation}
        Our current approach is constrained by the availability of sparse 2D supervisions derived from LiDAR point projections. 
        It is apparent that dense 2D supervisions could enhance the effectiveness of the regulatory stage during the training process of Vampire. 
        Furthermore, apart from the six camera views and the bird's eye view provided within a single frame, incorporating supervisory signals from additional perspectives (such as sequential frames) holds potential for improving our method. 
        This direction presents a compelling avenue for future research.

    \subsection{Board impact.}\label{ssec:board}
        While our primary focus is on academic research, our work bears the potential to significantly enhance road safety through the improvement of 3D environmental perception. 

        To mitigate the risk of malicious applications of our research in real-world scenarios, it is imperative that we exercise caution and implement stringent measures to safeguard sensitive data during the training and inference phases.
       
    \subsection{Code.}\label{ssec:code}
        In the supplementary material, we provide the code pertaining to our core design elements. 
        Our intention is to make the entire codebase and pre-trained models publicly accessible upon the publication of our paper.

\section{Additional Quantitative Results}\label{sec:additional}
    \subsection{Detailed results of 3D occupancy prediction.}\label{ssec:occ-full}
        Table \ref{tab:occ-full} presents a comprehensive performance comparison across 17 categories on the Occ3D-nuScenes dataset~\cite{tian2023occ3d}.
        Our method demonstrates state-of-the-art performance in the voxel-wise prediction task. Notably, Vampire exhibits a significant performance advantage over MonoScene~\cite{cao2022monoscene} and BEVDet~\cite{huang2021bevdet}.
        We also conducted comparisons with BEVFormer~\cite{li2022bevformer} and TPVFormer~\cite{huang2023tri}, where our model surpasses them by a margin of 1.45 and 0.5 in mIoU, respectively. 
        In the case of CTF-Occ~\cite{tian2023occ3d}, our achieved mIoU, while slightly lower at 28.33 compared to 28.53, is considered acceptable. 
        This difference can be attributed to our use of a smaller backbone (ResNet-50) and a lower input resolution.
        
    \subsection{Detailed results of LiDAR segmentation.}\label{ssec:lidar-seg-full}
        In Table \ref{tab:lidar-seg-full}, we provide a comprehensive breakdown of our LiDAR segmentation results. 
        Notably, when given camera images as input, Vampire surpasses the state-of-the-art image-based LiDAR segmentation method, TPVFormer~\cite{huang2023tri}, when employing the same backbone, with a notable margin of 7.1 points in mIoU.
        Remarkably, even in the absence of direct 3D LiDAR supervision, Vampire manages to achieve a higher mIoU than OccNet~\cite{sima2023_occnet} (62.2 compared to 53.0). Furthermore, Vampire demonstrates superior performance over RangeNet++~\cite{milioto2019rangenet++} in terms of IoU for most classes. 
        It is noteworthy that with additional training epochs, Vampire\textsuperscript{$\ddagger$} attains further performance improvements. 
        In addition, we report Vampire's performance on the nuScenes test set. 
        We posit that the performance potential of Vampire has not yet been fully realized, primarily due to hardware constraints.
    
    \subsection{Detailed results of 3D object detection.}\label{ssec:det-full}
        In Table \ref{tab:det-full}, we present comprehensive results for 3D object detection. For the baseline 3D object detection methods, we obtained the values for BEVFormer from the model repository of their \href{https://github.com/fundamentalvision/BEVFormer}{official repository}, using the [BEVFormer-tiny\_fp16] setting. 
        The values for BEVDet and BEVDepth are referenced from the BEVDepth paper~\cite{li2022bevdepth}.
        As detailed in our main text, Vampire exhibits its superiority in capturing intricate geometry details within 3D environments, as reflected by the mAP, mATE, and mASE metrics. These metrics are closely tied to the shape and geometry intricacies of the detected objects.
        Notably, the most pronounced distinction lies in the mAVE metric, where Vampire displays lower sensitivity to object velocity. 
        We attribute this phenomenon to the absence of additional occupancy flow (velocity information) annotations, which are employed by the joint training baseline methods (BEVNet, VoxNet, OccNet).
        
    \subsection{Ablation of position information.} \label{ssec:ablation-pos}
        To further investigate the significance of position information, we conducted additional ablation experiments. 
        In Table \ref{tab:ablation-pos}, we removed the injection of 3D normalized position coordinates based on our primary ablation study, full regulation setting. 
        We observed a performance degradation across all three tasks, underscoring the effectiveness of incorporating position information.
    
\begin{table}[ht]
	\centering
	\begin{tabular}{c | c c c}
		\toprule
	    Cat. Pos. & Occ.$\uparrow$ & Seg.$\uparrow$ & Det.$\uparrow$ \\
		\midrule
          & 25.0 & 61.5 & 0.311 \\
        \checkmark & \bf 25.8 &\bf 62.6 &\bf 0.318\\
		\bottomrule
	\end{tabular}
 \vspace{-1.5mm}
 \caption{\textbf{Ablation study for position information.}
 ``Cat. Pos.'' stands for concatenating normalized 3D position coordinates.
  ``Occ.'' represents 3D occupancy prediction, ``Seg.'' refers to LiDAR segmentation, ``Det.'' denotes 3D object detection.
  The table provides a comprehensive view of the effectiveness of position information.
	}
	\label{tab:ablation-pos}
 \vspace{-2mm}
\end{table}
    \subsection{Training resource comparison.}\label{ssec:train-resouce}
        We present a comparison of the training resources required by our Vampire method and existing state-of-the-art approaches. 
        Vampire demonstrates a moderate demand for GPU resources during training, rendering our approach more practical and accessible to the wider research community.

\begin{table}[h]
	\centering
	\begin{tabular}{l | c | c | c}
		\toprule
	    Method & Device & Memory & Ref. \\
		\midrule
        TPVFormer~\shortcite{huang2023tri} & RTX3090 & 23G & \href{https://github.com/wzzheng/TPVFormer/issues/10}{issue}\\
        TPVFormer~\shortcite{huang2023tri} & A100& 39G & \href{https://github.com/wzzheng/TPVFormer/issues/10}{issue} \\
        OccNet-tiny~\shortcite{sima2023_occnet} & V100/RTX3090& 20G & \href{https://github.com/OpenDriveLab/OccNet/issues/7}{issue}\\
        OccNet-base~\shortcite{sima2023_occnet} & A100& 60G & \href{https://github.com/OpenDriveLab/OccNet/issues/7}{issue}\\
        \midrule
        \bf Vampire (ours) & RTX3080Ti &10.6G & - \\
		\bottomrule
	\end{tabular}
 \vspace{-1.5mm}
 \caption{\textbf{Training resource comparison.}
    Vampire necessitates only moderate GPU resources for training, making it a more practical choice for the broader research community.
	}
	\label{tab:resource}
 \vspace{-2mm}
\end{table}

\section{Additional Qualitative Results}\label{sec:qualitative}
    \subsection{Detection results on nuScenes.}\label{ssec:qualitative-nuscenes}
\begin{figure*}[hb]
  \centering
  \includegraphics[trim=4 4 4 4, clip,height=22cm]{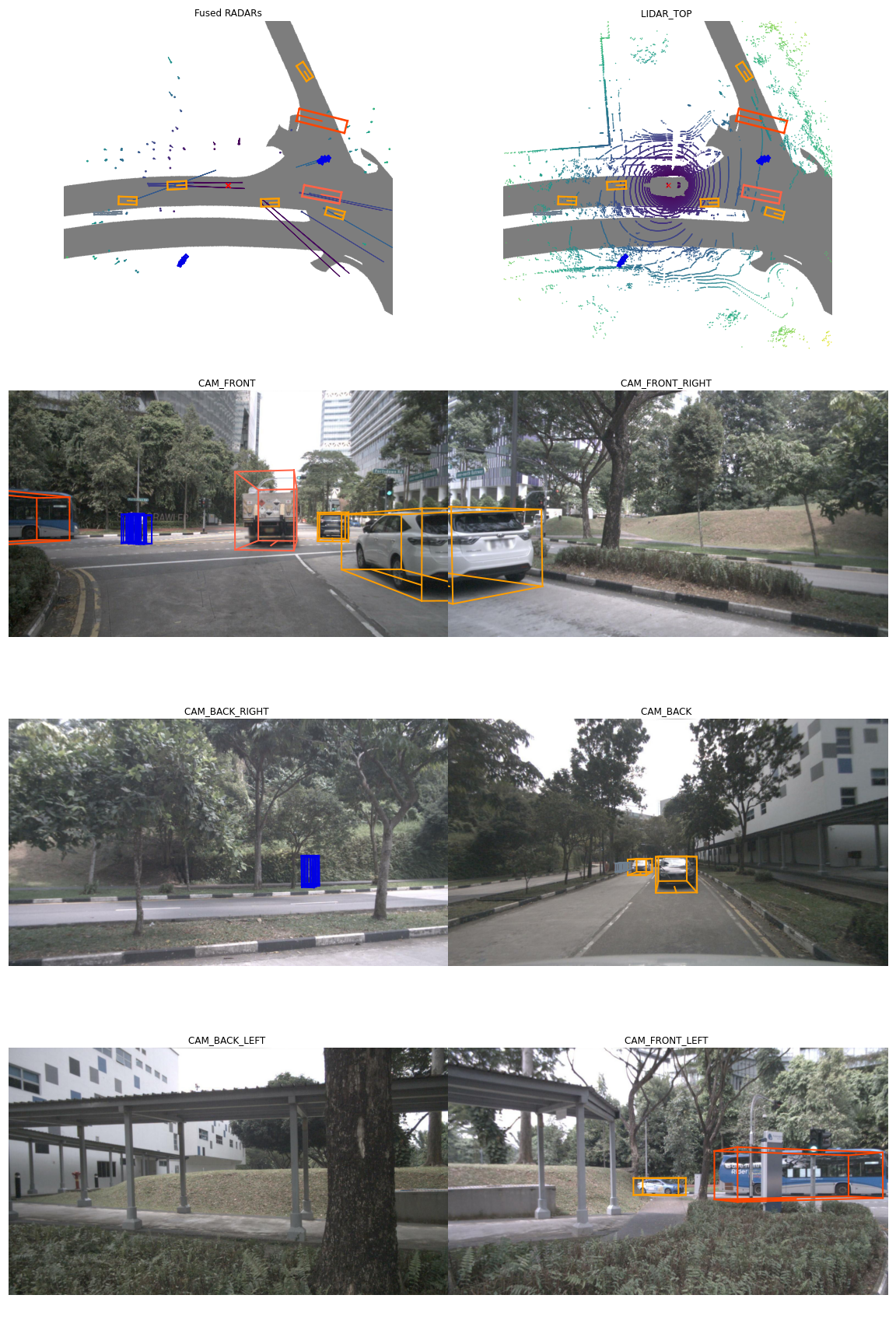}
  \caption{\textbf{Visualization for 3D object detection (1).} 
  Our results indicate that Vampire excels in detecting large objects with rigid geometries. However, it exhibits limitations in accurately detecting pedestrians.
  }
  \label{supply-fig:det-vis1}
\end{figure*}

\begin{figure*}[hb]
  \centering
  \includegraphics[trim=4 4 4 4, clip,height=22cm]{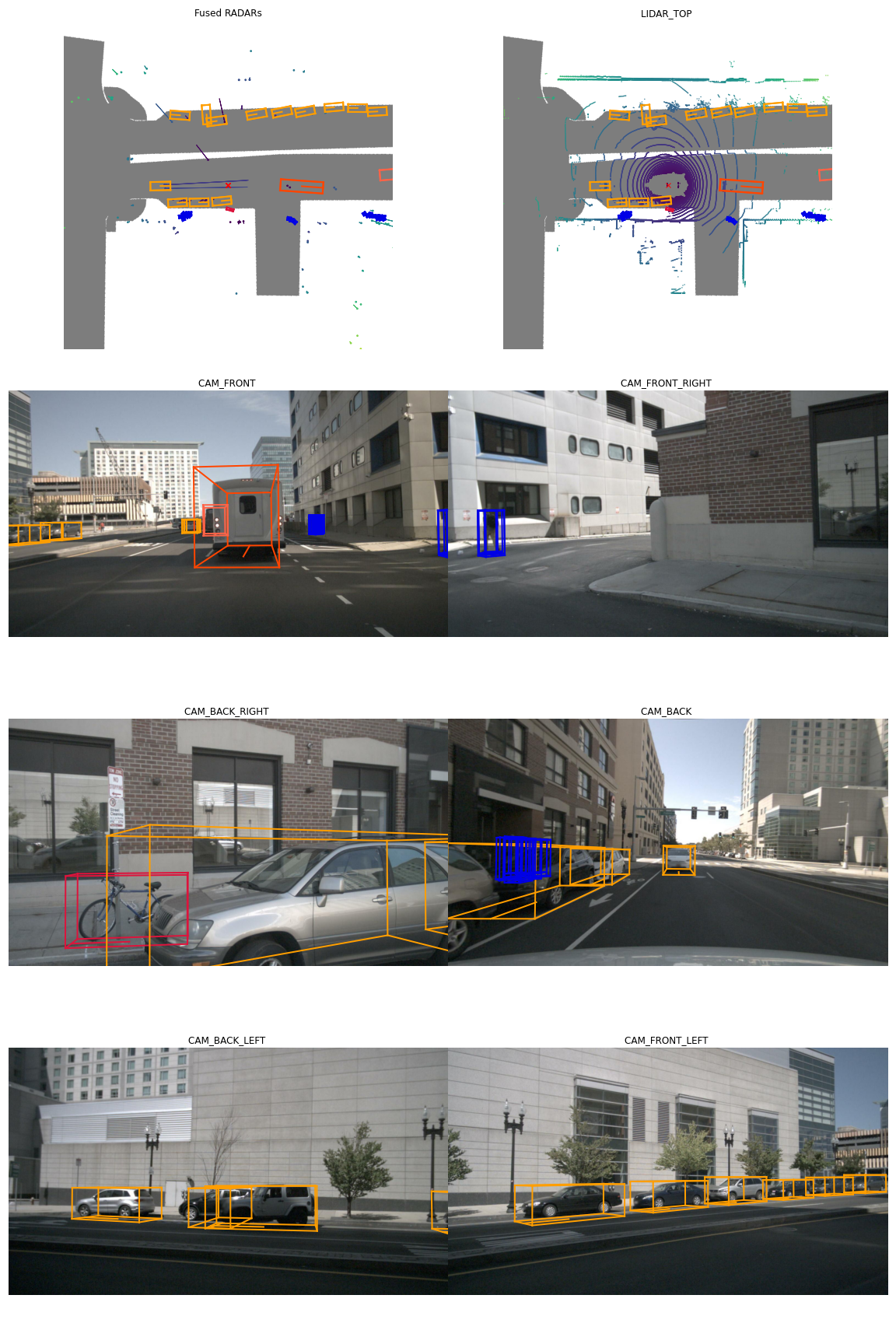}
  \caption{\textbf{Visualization for 3D object detection (2).} 
  Our results indicate that Vampire excels in detecting large objects with rigid geometries. However, it exhibits limitations in accurately detecting pedestrians.
  }
  \label{supply-fig:det-vis2}
\end{figure*}

        In Figures \ref{supply-fig:det-vis1} and \ref{supply-fig:det-vis2}, we present qualitative examples of 3D detection on the nuScenes validation set~\cite{caesar2020nuscenes}. 
        Our method demonstrates robust performance in detecting larger categories such as cars and buses. 
        However, it shows comparatively lower performance in detecting pedestrians. 
        This limitation may be attributed to the voxel size, as 0.4 meters might be too large to effectively sense pedestrians.
        
    \subsection{Rendered results and 3D visualizations.}\label{ssec:rendered-results}
\begin{figure*}
  \centering
  \includegraphics[trim=4 4 4 4, clip,width=\linewidth]{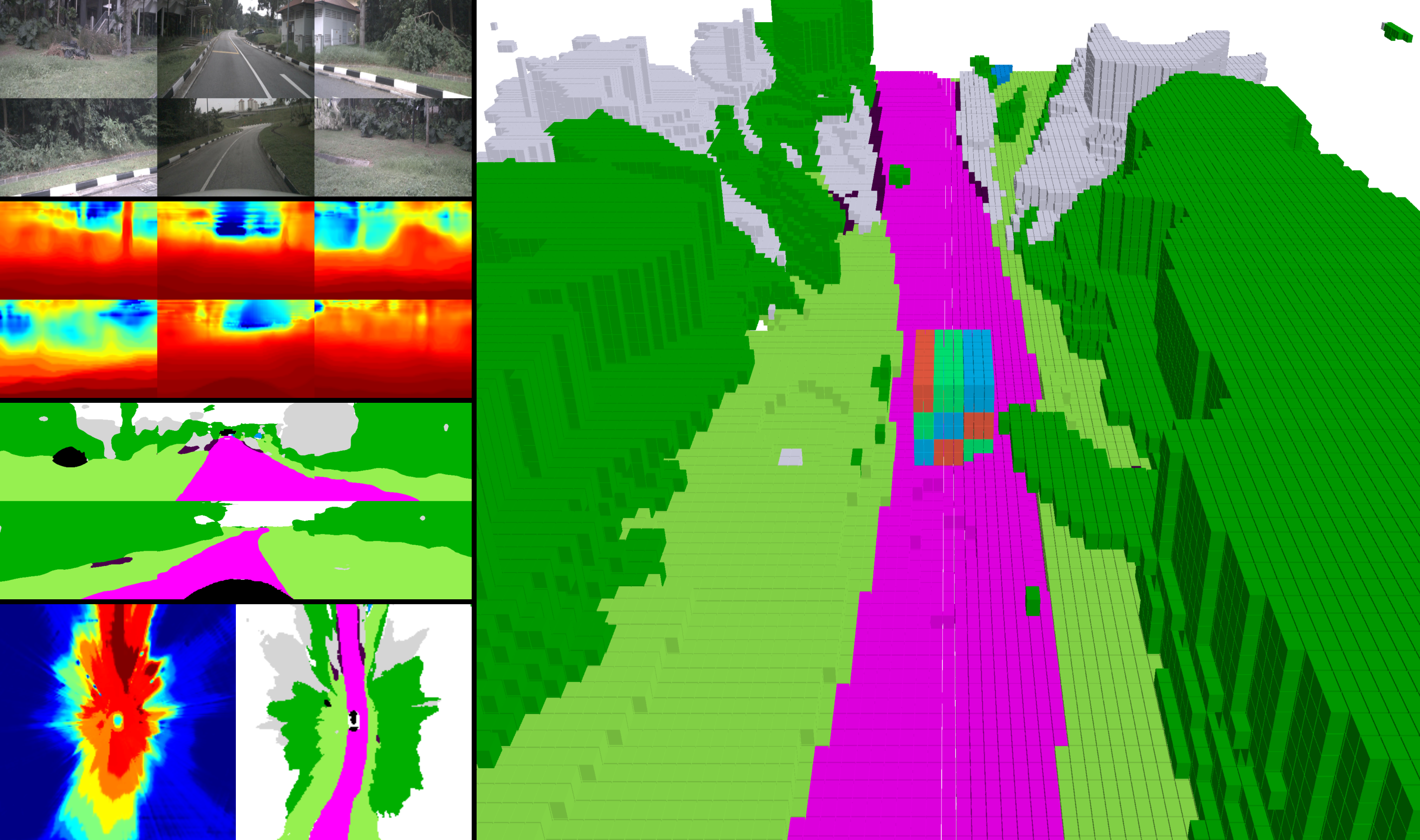}
  \includegraphics[trim=4 4 4 4, clip,width=\linewidth]{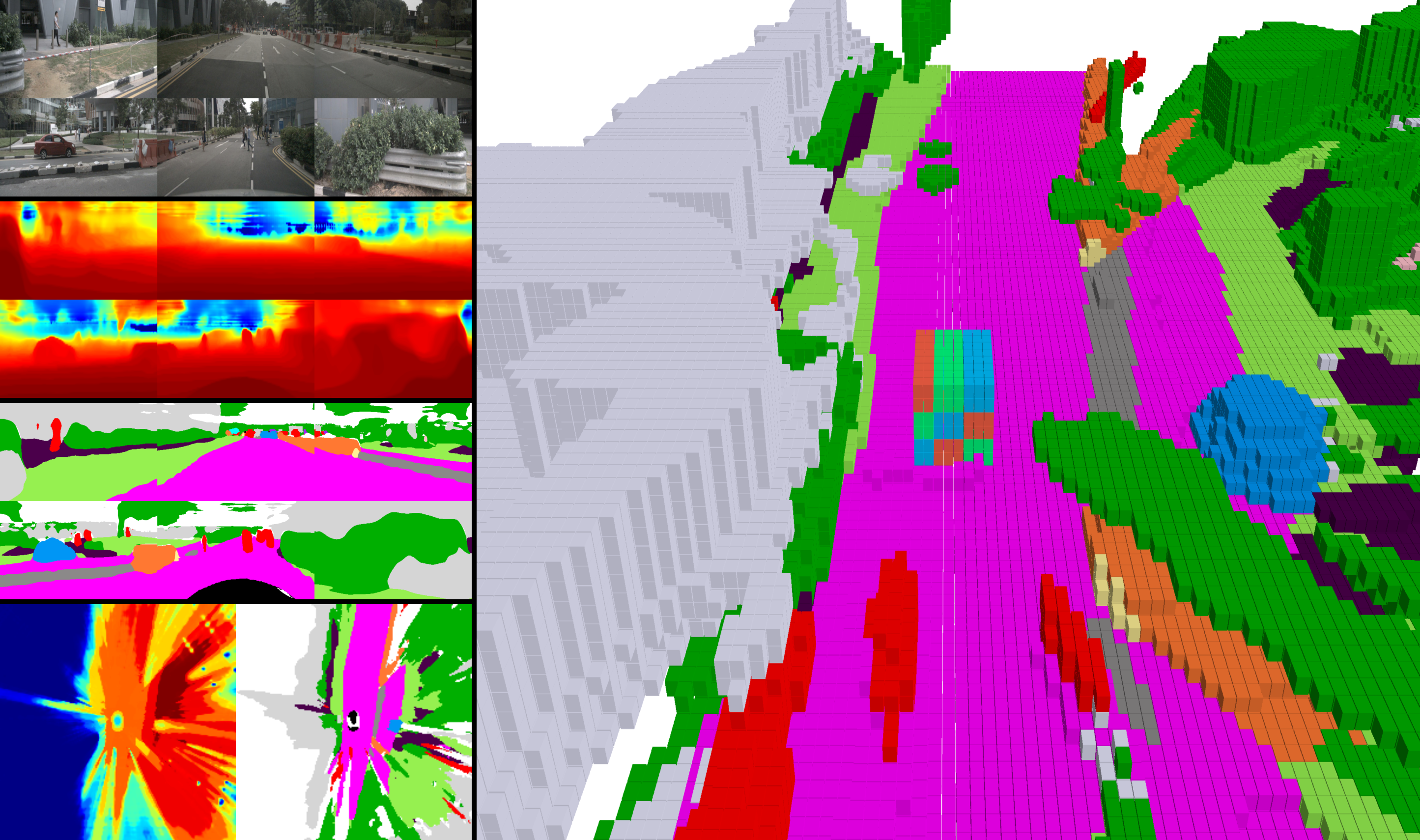}
  \caption{\textbf{Rendered results and 3D occupancy visualization (1)}.
  }
  \label{supply-vis:1}
\end{figure*}

\begin{figure*}
  \centering
  \includegraphics[trim=4 4 4 4, clip,width=\linewidth]{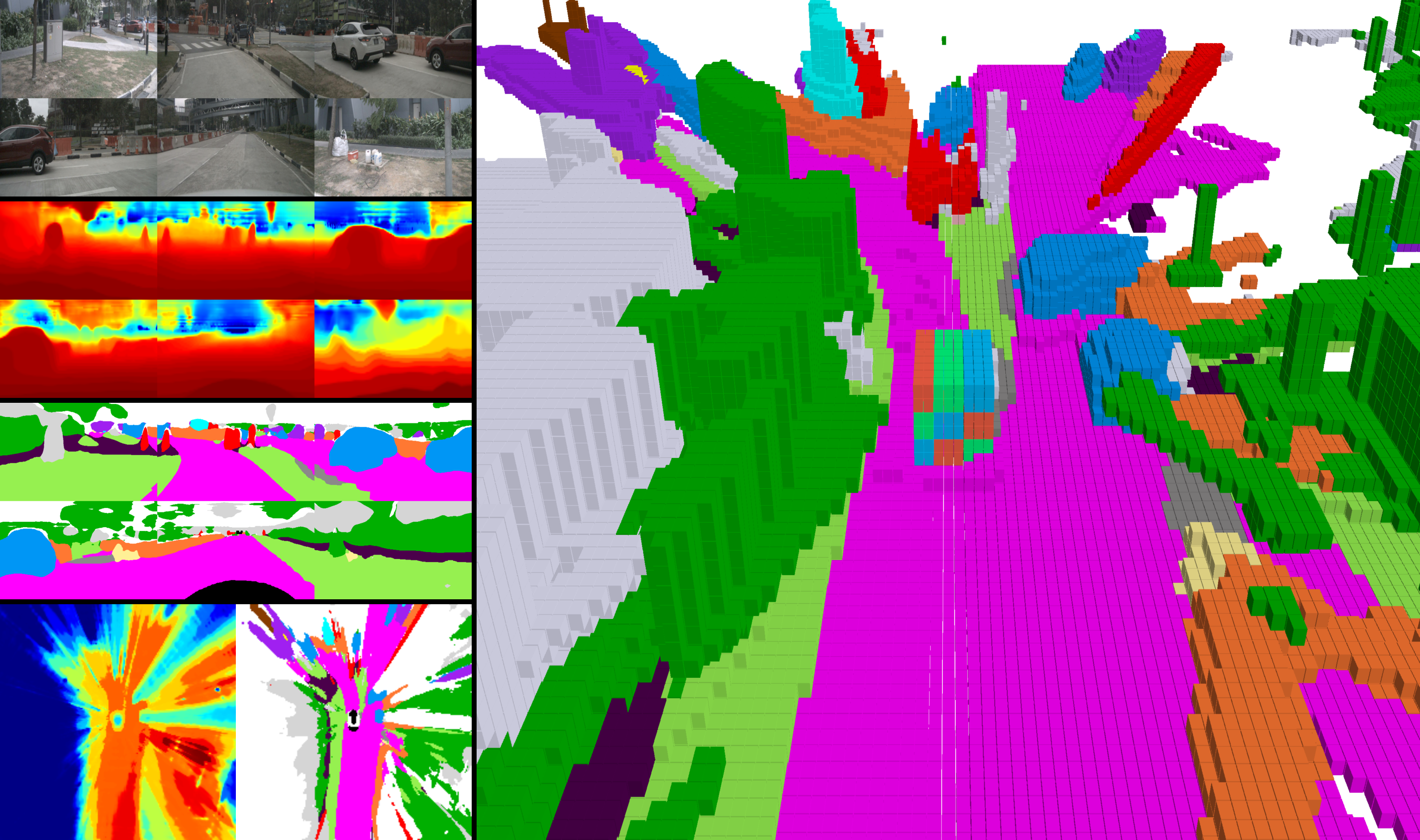}
  \includegraphics[trim=4 4 4 4, clip,width=\linewidth]{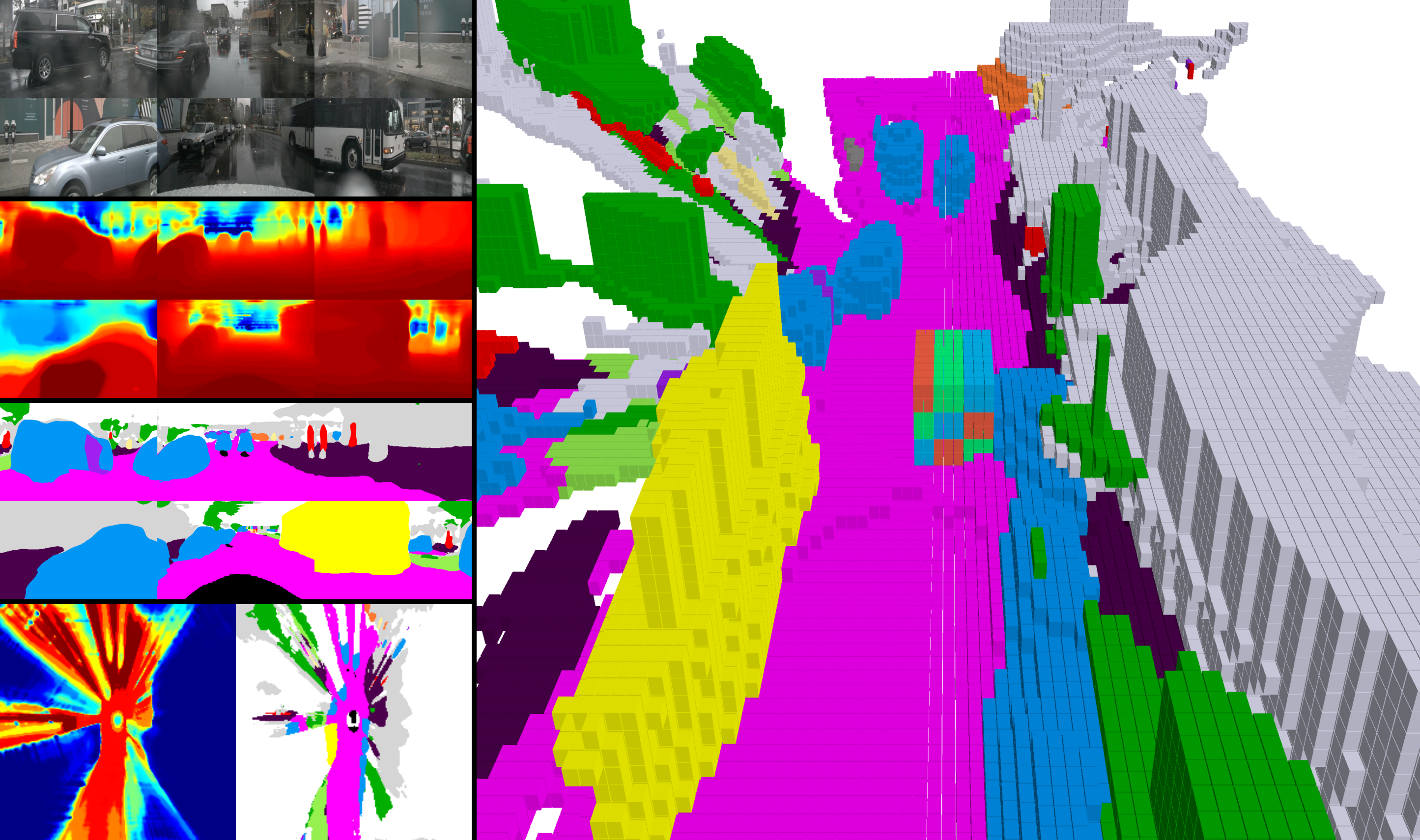}
  \caption{\textbf{Rendered results and 3D occupancy visualization (2).}
  }
  \label{supply-vis:2}
\end{figure*}

        We present rendered results in Figures \ref{supply-vis:1} and \ref{supply-vis:2}, selected from our demonstration videos.

    \subsection{Occupancy demo video.}\label{ssec:occupancy-results}
        For our occupancy demonstration video, we utilized scene sequences extracted from the original nuScenes validation set~\cite{caesar2020nuscenes}, specifically, the continuous scene sequences spanning from 0012 to 0018 and from 0904 to 0917.
        Our 3D visualizations were implemented using Mayavi~\cite{ramachandran2011mayavi} and were based on the implementations of TPVFormer~\cite{huang2023tri} and OccFormer~\cite{zhang2023occformer}. 
        Within each frame of the video, we showcase the input images, rendered depth maps, rendered semantic maps from both camera and bird's eye views. 
        Additionally, we provide an accumulated density Bird's Eye View (BEV) map.
        The generated video clips, labeled as \textit{scene-0012-0018.mp4} and \textit{scene-0904-0917.mp4}, are available within the supplementary zip file.

\definecolor{nothers}{RGB}{0, 0, 0}
\definecolor{nbarrier}{RGB}{255, 120, 50}
\definecolor{nbicycle}{RGB}{255, 192, 203}
\definecolor{nbus}{RGB}{255, 255, 0}
\definecolor{ncar}{RGB}{0, 150, 245}
\definecolor{nconstruct}{RGB}{0, 255, 255}
\definecolor{nmotor}{RGB}{200, 180, 0}
\definecolor{npedestrian}{RGB}{255, 0, 0}
\definecolor{ntraffic}{RGB}{255, 240, 150}
\definecolor{ntrailer}{RGB}{135, 60, 0}
\definecolor{ntruck}{RGB}{160, 32, 240}
\definecolor{ndriveable}{RGB}{255, 0, 255}
\definecolor{nother}{RGB}{139, 137, 137}
\definecolor{nsidewalk}{RGB}{75, 0, 75}
\definecolor{nterrain}{RGB}{150, 240, 80}
\definecolor{nmanmade}{RGB}{213, 213, 213}
\definecolor{nvegetation}{RGB}{0, 175, 0}
\begin{table*}[!ht]
	\scriptsize
	\setlength\tabcolsep{1.8pt}
	\centering
	\begin{tabular}{l|c|c|c c c c c c c c c c c c c c c c c|c}
		\toprule
		Method
		& Backbone 
        & \makecell{Image \\ Size}
        & \rotatebox{90}{\textcolor{nothers}{$\blacksquare$} Others}
		& \rotatebox{90}{\textcolor{nbarrier}{$\blacksquare$} Barrier}
		
		& \rotatebox{90}{\textcolor{nbicycle}{$\blacksquare$} Bicycle}
		
		& \rotatebox{90}{\textcolor{nbus}{$\blacksquare$} Bus}

		& \rotatebox{90}{\textcolor{ncar}{$\blacksquare$} Car}

		& \rotatebox{90}{\textcolor{nconstruct}{$\blacksquare$} Const. Veh.}

		& \rotatebox{90}{\textcolor{nmotor}{$\blacksquare$} Motorcycle}

		& \rotatebox{90}{\textcolor{npedestrian}{$\blacksquare$} Pedestrian}

		& \rotatebox{90}{\textcolor{ntraffic}{$\blacksquare$} Traffic cone}

		& \rotatebox{90}{\textcolor{ntrailer}{$\blacksquare$} Trailer}

		& \rotatebox{90}{\textcolor{ntruck}{$\blacksquare$} Truck}

		& \rotatebox{90}{\textcolor{ndriveable}{$\blacksquare$} Dri. Suf.}

		& \rotatebox{90}{\textcolor{nother}{$\blacksquare$} Other flat}

		& \rotatebox{90}{\textcolor{nsidewalk}{$\blacksquare$} Sidewalk}

		& \rotatebox{90}{\textcolor{nterrain}{$\blacksquare$} Terrain}

		& \rotatebox{90}{\textcolor{nmanmade}{$\blacksquare$} Manmade}

		& \rotatebox{90}{\textcolor{nvegetation}{$\blacksquare$} Vegetation}
  
        & mIoU
		\\
		\midrule
        MonoScene~\shortcite{tian2023occ3d}&Effi.B7 & {900$\times$1600} &1.75 &7.23 &4.26 &4.93 &9.38 &5.67 &3.98 &3.01 &5.90 &4.45 & 7.17 & 14.91 &6.32 &7.92 &7.43 &1.01 &7.65 &6.06 \\
        \midrule
        BEVDet~\shortcite{tian2023occ3d} &\multirow{5}{*}{R101} & \multirow{5}{*}{900$\times$1600} &4.39 &30.31 &0.23 &32.26 &34.47 &12.97 &10.34 &10.36 &6.26 &8.93 &23.65 &52.27 &24.61 &26.06 &22.31 &15.04 &15.10 &19.38\\
        OccFormer~\shortcite{tian2023occ3d} & & &5.94 &30.29 &12.32 &34.40 &39.17 &14.44 &16.45 &17.22 &9.27 &13.90 &26.36 &50.99 &30.96 &34.66 &22.73 &6.76 &6.97 &21.93 \\
        BEVFormer~\shortcite{tian2023occ3d} & & &5.85 &37.83 &{17.87} &{40.44} &{42.43} &7.36 &{23.88} &{21.81} &{20.98} &22.38 &30.70 &55.35 &28.36 &36.00 &28.06 &20.04 &17.69 &26.88 \\
        TPVFormer~\shortcite{tian2023occ3d} & & &7.22 &{38.90} &13.67 &\bf 40.78 &\bf 45.90 &\bf 17.23 &19.99 &18.85 &14.30 &\bf 26.69 &\bf 34.17 &{55.65} &\bf 35.47 &37.55 &30.70 &19.40 &16.78 &27.83 \\
        CTF-Occ~\shortcite{tian2023occ3d} & & &\bf 8.09 &\bf 39.33 &\bf 20.56 &38.29 &42.24 &{16.93} &\bf 24.52 &\bf 22.72 &\bf 21.05 &{22.98} &{31.11} &53.33 &{33.84} &{37.98} &{33.23} &{20.79} &{18.00} &\bf 28.53 \\
        \midrule
        \textbf{Vampire} (ours) &R50 &{256$\times$704} &{7.48} &32.64 &16.15 &36.73 &41.44 &16.59 &20.64 &16.55 &15.09 &21.02 &28.47 &\bf 67.96 &33.73 &\bf 41.61 &\bf 40.76 &\bf 24.53 &\bf 20.26 & {28.33}\\
		\bottomrule
	\end{tabular}
     \caption{\textbf{Detailed 3D occupancy prediction results on Occ3D-nuScenes~\shortcite{tian2023occ3d}.} 
     ``Effi.NetB7'' stands for EfficientNetB7,
     ``Const. Veh'' represents construction vehicle and ``Dri.Sur.'' is for driveable surface.
     Despite image backbone and input size differences, Vampire achieves comparable performance with state-of-the-art methods. }
	\label{tab:occ-full}
\end{table*}
\definecolor{nbarrier}{RGB}{255, 120, 50}
\definecolor{nbicycle}{RGB}{255, 192, 203}
\definecolor{nbus}{RGB}{255, 255, 0}
\definecolor{ncar}{RGB}{0, 150, 245}
\definecolor{nconstruct}{RGB}{0, 255, 255}
\definecolor{nmotor}{RGB}{200, 180, 0}
\definecolor{npedestrian}{RGB}{255, 0, 0}
\definecolor{ntraffic}{RGB}{255, 240, 150}
\definecolor{ntrailer}{RGB}{135, 60, 0}
\definecolor{ntruck}{RGB}{160, 32, 240}
\definecolor{ndriveable}{RGB}{255, 0, 255}
\definecolor{nother}{RGB}{139, 137, 137}
\definecolor{nsidewalk}{RGB}{75, 0, 75}
\definecolor{nterrain}{RGB}{150, 240, 80}
\definecolor{nmanmade}{RGB}{213, 213, 213}
\definecolor{nvegetation}{RGB}{0, 175, 0}
\begin{table*}[!ht]
	\scriptsize
	\setlength\tabcolsep{2pt}
	\centering
	\begin{tabular}{l|c|c|c|c c c c c c c c c c c c c c c c |c}
		\toprule
		Method
		& \makecell{Input \\ Modality} 
        & Backbone & \makecell{Image \\ Size} 
		& \rotatebox{90}{\textcolor{nbarrier}{$\blacksquare$} Barrier}
		
		& \rotatebox{90}{\textcolor{nbicycle}{$\blacksquare$} Bicycle}
		
		& \rotatebox{90}{\textcolor{nbus}{$\blacksquare$} Bus}

		& \rotatebox{90}{\textcolor{ncar}{$\blacksquare$} Car}

		& \rotatebox{90}{\textcolor{nconstruct}{$\blacksquare$} Const. Veh.}

		& \rotatebox{90}{\textcolor{nmotor}{$\blacksquare$} Motorcycle}

		& \rotatebox{90}{\textcolor{npedestrian}{$\blacksquare$} Pedestrian}

		& \rotatebox{90}{\textcolor{ntraffic}{$\blacksquare$} Traffic cone}

		& \rotatebox{90}{\textcolor{ntrailer}{$\blacksquare$} Trailer}

		& \rotatebox{90}{\textcolor{ntruck}{$\blacksquare$} Truck}

		& \rotatebox{90}{\textcolor{ndriveable}{$\blacksquare$} Dri. Suf.}

		& \rotatebox{90}{\textcolor{nother}{$\blacksquare$} Other flat}

		& \rotatebox{90}{\textcolor{nsidewalk}{$\blacksquare$} Sidewalk}

		& \rotatebox{90}{\textcolor{nterrain}{$\blacksquare$} Terrain}

		& \rotatebox{90}{\textcolor{nmanmade}{$\blacksquare$} Manmade}

		& \rotatebox{90}{\textcolor{nvegetation}{$\blacksquare$} Vegetation}

        & mIoU
		\\
		\hline
        \multicolumn{21}{c}{nuScenes \textit{validation} set}\\
        \hline
        RangeNet++~\shortcite{milioto2019rangenet++} & \multirow{2}{*}{LiDAR} & \multirow{2}{*}{-} & \multirow{2}{*}{-} & 66.0 & 21.3 & 77.2 & 80.9 & 30.2 & 66.8 & 69.6 &  52.1 & 54.2 & {72.3} & {94.1} & 66.6 & 63.5 & 70.1 & 83.1 & 79.8 &  65.5\\
    
    
    
        Cylinder3D++~\shortcite{zhu2021cylindrical} &  & & & {76.4} & {40.3} & {91.2} & {93.8} & {51.3} & {78.0} & {78.9} & {64.9} & {62.1} & {84.4} & {96.8} & {71.6} & {76.4} & {75.4} & {90.5} & {87.4} & {76.1} \\
        \midrule	
        BEVFormer~\shortcite{li2022bevformer} & \multirow{4}{*}{Camera} & \multirow{4}{*}{\makecell{R101}} &\multirow{4}{*}{900$\times$1600} & 54.0 & 22.8 & 76.7 & 74.0 & 45.8 & 53.1 & 44.5 & 24.7 & 54.7 & 65.5 & 88.5 & 58.1 & 50.5 & 52.8 & 71.0 & 63.0  & 56.2\\
        TPVFormer~\shortcite{huang2023tri} & & &  & 70.0 & \bf 40.9 & \bf 93.7 & 85.6 & 49.8 & \bf 68.4 & \bf 59.7 & 38.2 & \bf 65.3 & \bf 83.0 & 93.3 & 64.4 & 64.3 & 64.5 & \bf 81.6 & \bf 79.3 & \bf 68.9 \\ 
        TPVFormer\textsuperscript{$\dagger$}~\shortcite{sima2023_occnet} &  &  & &66.0 &24.5 &80.9 &74.3 &47.0 &47.1 &33.4 &14.5 &54.0 &70.8 &88.6 &61.6 &59.5 &63.2 &75.8 &74.2 &58.5 \\
        OccNet\textsuperscript{$\dagger$}~\shortcite{sima2023_occnet} &  &  & &67.0 &32.6 &77.4 &73.9 &37.6 &50.9 &51.5 &33.7 &52.2 &67.1 &88.7 &58.0 &58.0 &63.1 &78.9 &77.0 &60.5 \\
        \midrule
        TPVFormer~\shortcite{huang2023tri} & \multirow{5}{*}{Camera} & \multirow{5}{*}{R50} & 450$\times$800 &64.9 &27.0 &83.0& 82.8 &38.3 &27.4 &44.9 &24.0 &55.4 &73.6 &91.7 &60.7 &59.8 &61.1 &78.2 &76.5& 59.3 \\
        OccNet\textsuperscript{$\dagger$}~\shortcite{sima2023_occnet} & & & 900$\times$1600 &65.9 &22.8 &64.1 &72.7 &32.7 &28.7 &52.2 &17.6 &22.1 &51.3 &89.1 &57.4 &58.1 &64.3 &75.1 &73.9& 53.0 \\
        \bf Vampire (ours) & & & {256$\times$704} &\bf 73.1 &35.8 &90.0 &\bf 87.0 &48.7 &42.1 &48.5 &\bf 39.7 &60.1 &77.9 &94.7 & 70.5 &68.2 &\bf 68.4 &80.4 &76.9 & 66.4 \\
        \bf Vampire (ours) \textsuperscript{$\dagger$} & & & {256$\times$704} &69.1 &29.0 &85.6 &79.8 &44.0 &46.0 &44.7 &29.0 &53.8 &74.9 &91.6 &66.6 &64.7 &66.0 &77.3 &72.9 &62.2 \\
        \bf Vampire (ours) \textsuperscript{$\ddagger$} & & & {256$\times$704} &72.6 &32.6 &91.8 &85.3 &\bf 50.1 &62.9 &57.3 &38.3 &62.4 &79.1 &\bf 95.0 &\bf 71.6 &\bf 68.3 &68.2 & 80.5 &77.6 &68.3\\
        \hline
        \multicolumn{21}{c}{nuScenes \textit{test} set}\\
        \hline
         TPVFormer~\shortcite{huang2023tri} & \multirow{2}{*}{Camera}& \multirow{2}{*}{R101}& \multirow{2}{*}{900$\times$1600} & 74.0 &27.5 &86.3 &85.5 &\bf 60.7 &68.0 &62.1 &49.1 &81.9 &\bf 68.4 &94.1 &59.5 &66.5 &63.5 &83.8 &79.9 &69.4\\
         OccFormer~\shortcite{zhang2023occformer} & & & &72.8 &\bf 29.9 &\bf 87.9 &85.6 &57.1 &\bf 74.9 &\bf 63.2 &\bf 53.4 &\bf 83.0 &67.6 &94.8 &61.9 &70.0 &\bf 66.0 &\bf 84.0 &\bf 80.5 &\bf 70.8\\
         \midrule
         TPVFormer~\shortcite{huang2023tri} & \multirow{2}{*}{Camera}& \multirow{2}{*}{R50}& 450$\times$800 &65.6 &15.7 &75.1 &80.0 &45.8 &43.1 &44.3 &26.8 &72.8 &55.9 &92.3 &53.7 &61.0 &59.2 &79.7 &75.6 &59.2\\
         \bf Vampire (ours) \textsuperscript{$\ddagger$} & & & {256$\times$704} &\bf 75.4 &26.0 &79.3 &\bf 87.2 &53.1 &68.8 &49.7 &46.3 &79.1 &67.1 &\bf 95.6 &\bf 64.2 &\bf 71.0 &64.3 &81.8 &77.2 &67.9 \\
        \bottomrule

	\end{tabular}
 \caption{\textbf{Detailed LiDAR segmentation results on nuScenes val set and test set.} 
    ``Const. Veh'' represents construction vehicle and ``Dri.Sur.'' is for driveable surface.
    Mark $\dagger$ indicates methods trained without direct LiDAR supervision but only occupancy semantic labels. 
    Mark $\ddagger$ indicates methods trained with additional epochs, we train the Vampire \textsuperscript{$\ddagger$} for 48 epochs.
    Despite image backbone and input size differences, our Vampire achieves comparable performance with state-of-the-art image-based methods and even LiDAR-based method.
	}
	\label{tab:lidar-seg-full}
\end{table*} 
\begin{table*}[!h]
	\setlength\tabcolsep{1.5pt}
	\centering
	\begin{tabular}{l|c|c c| c c c c c}
		\toprule
	    Method & Joint. & mAP$\uparrow$ & NDS $\uparrow$& mATE $\downarrow$ & mASE $\downarrow$ &mAOE $\downarrow$ & mAVE $\downarrow$ & mAAE $\downarrow$\\
        \midrule
        BEVFormer~\shortcite{li2022bevformer}  &  & 0.257 & 0.359 &0.884 &0.290 &0.626 &0.673 &0.225\\
        BEVDet~\shortcite{li2022bevdepth}   & & 0.286 &0.372 & - & - & - & - & - \\
        BEVDetph~\shortcite{li2022bevdepth}   &  & 0.322 &0.367 &0.707 & - &0.636 & - & - \\
        \midrule
        BEVNet~\shortcite{sima2023_occnet}  & \multirow{4}{*}{\checkmark} &0.271 &\bf 0.390 &0.835 &0.293 &\bf 0.578 &\bf 0.541 & 0.211\\
        VoxNet~\shortcite{sima2023_occnet}   &  &0.277 & 0.387 &0.828 &0.285 &0.586 &0.614 &0.203\\
        OccNet~\shortcite{sima2023_occnet}    & &0.276 & \bf 0.390 &0.842 &0.292 &0.585 &0.570 &\bf 0.190\\
        \bf Vampire (ours)   & & \bf 0.301 & 0.354 &\bf 0.741 &\bf 0.278 &0.649 &1.043 &0.297\\
		\bottomrule
	\end{tabular}
    \caption{
      \textbf{Detailed 3D object detection results on nuScenes val set.}
    	}
	\label{tab:det-full}
\end{table*}


\end{document}